\documentclass[11pt]{article}

\usepackage[preprint]{acl}
\usepackage{times}
\usepackage{latexsym}

\usepackage[T1]{fontenc}

\usepackage[utf8]{inputenc}

\usepackage{microtype}
\usepackage{pifont}

\usepackage{inconsolata}

\usepackage{graphicx}
\usepackage{times}
\usepackage{latexsym}
\usepackage{graphicx}
\usepackage{subfigure}
\usepackage{booktabs} 

\usepackage{multirow}
\usepackage{verbatim}
\usepackage{graphicx}
\usepackage{pifont}
\usepackage{adjustbox}
\usepackage{enumitem}
\usepackage{amsfonts}
\usepackage{array}
\usepackage{caption}
\usepackage{color}
\usepackage{xcolor}
\usepackage{amssymb}
\usepackage{mathtools}
\usepackage{amsthm}
\usepackage{colortbl}
\usepackage{microtype}
\usepackage{amssymb}   
\usepackage{textcomp}  

\title{GenProve: Learning to Generate Text with Fine-Grained Provenance}

\usepackage{amssymb} 

\DeclareRobustCommand{\AffA}{2}
\DeclareRobustCommand{\AffB}{3}
\DeclareRobustCommand{\AffC}{4}
\DeclareRobustCommand{\AffD}{1}

\DeclareRobustCommand{\EqC}{\textdagger}
\DeclareRobustCommand{\Corr}{*}

\author{
  Jingxuan Wei\textsuperscript{\AffD,\AffA,\AffC,\EqC} \quad
  Xingyue Wang\textsuperscript{\AffA,\AffC,\EqC} \quad
  Yanghaoyu Liao\textsuperscript{\AffA,\AffC,\EqC} \quad
  Jie Dong\textsuperscript{\AffA,\AffC,\EqC} \\
  \textbf{Yuchen Liu}\textsuperscript{} \quad
  \textbf{Caijun Jia}\textsuperscript{\AffA,\AffC} \quad
  \textbf{Bihui Yu}\textsuperscript{\AffA,\AffC} \quad
  \textbf{Junnan Zhu}\textsuperscript{\AffB\Corr} \\
  \textsuperscript{\AffD}\ Key Laboratory of Computing Power Network and Information Security, Ministry of \\ Education,  Shandong  Computer Science Center (National Supercomputer Center \\ in Jinan),  Qilu  University of  Technology  (Shandong Academy of Sciences)  \\
  \textsuperscript{\AffA}\ Shenyang Institute of Computing Technology, Chinese Academy of Sciences \\
  \textsuperscript{\AffB}\ MAIS, Institute of Automation, Chinese Academy of Sciences \\
  \textsuperscript{\AffC}\ University of Chinese Academy of Sciences 
}

\PassOptionsToPackage{table,xcdraw,svgnames,dvipsnames}{xcolor}
\usepackage[table]{xcolor}
\usepackage[most]{tcolorbox}

\definecolor{promptcolor}{HTML}{F7E5DD}
\definecolor{promptcolorheader}{HTML}{D1C2BB}

\newtcolorbox{promptbox}[1][]{
  enhanced, breakable,
  top=0.3em,bottom=0.3em,left=0.5em,right=0.5em,
  toptitle=0.3em,bottomtitle=0.2em,boxsep=0pt,
  colframe=promptcolorheader,
  colback=promptcolor!70,   
  boxrule=0.5pt,
  width=\linewidth,
  title={\footnotesize #1}
}

\begin{document}
\maketitle
\begingroup
\renewcommand\thefootnote{}
\footnotetext{\EqC\ Equal contribution \quad \Corr\ Corresponding author}
\endgroup
\begin{abstract}
Large language models (LLM) often hallucinate, and while adding citations is a common solution, it is frequently insufficient for accountability as users struggle to verify how a cited source supports a generated claim. Existing methods are typically coarse-grained and fail to distinguish between direct quotes and complex reasoning. In this paper, we introduce \textbf{Generation-time Fine-grained Provenance}, a task where models must generate fluent answers while simultaneously producing structured, sentence-level provenance triples. To enable this, we present \textbf{ReFInE} (\textbf{Re}lation-aware \textbf{F}ine-grained \textbf{In}terpretability \& \textbf{E}vidence), a dataset featuring expert-verified annotations that distinguish between \textit{Quotation}, \textit{Compression}, and \textit{Inference}. Building on ReFInE, we propose \textbf{GenProve}, a framework that combines Supervised Fine-Tuning (SFT) with Group Relative Policy Optimization (GRPO). By optimizing a composite reward for answer fidelity and provenance correctness, GenProve significantly outperforms 14 strong LLMs in joint evaluation. Crucially, our analysis uncovers a reasoning gap where models excel at surface-level quotation but struggle significantly with inference-based provenance, suggesting that verifiable reasoning remains a frontier challenge distinct from surface-level citation. The code is publicly available at \url{https://github.com/yanghaoyuliao/GenProve}.
\end{abstract}

\section{Introduction}
While LLMs demonstrate impressive fluency, their tendency to hallucinate remains a major barrier to widespread adoption~\cite{fan2025trustworthiness}. Users need to verify not just whether an answer sounds correct, but exactly where it comes from in the external evidence. To address this issue, current systems typically use Retrieval-Augmented Generation (RAG) or simply add citations to the text~\cite{gao-2023-enabling-llms-citation, li-2024-citation-enhanced}. However, these standard approaches often function as \textit{black boxes} because they provide a list of documents yet fail to specify \emph{which} exact sentence supports a claim or \emph{how} that evidence is used. This ambiguity leaves users guessing whether the model is directly quoting a fact, summarizing scattered details, or making a logical inference, which makes verification difficult and inefficient. For rigorous verification, knowing \emph{how} a model uses evidence (e.g., inferring vs. quoting) is as important as knowing \emph{which} document it used.

\begin{figure}[t]
  \centering
  \includegraphics[width=\linewidth]{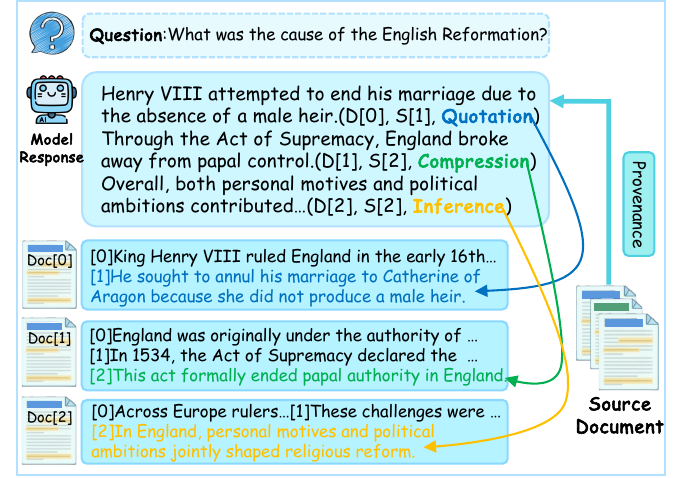}
  \caption{Overview of \textbf{Generation-time Fine-grained Provenance}. Given a query and source documents, the model simultaneously produces the answer and structured triples (DocID, SentID, Relation) to explain how the evidence supports each generated sentence.}
  \label{fig:intro}
\end{figure}

We advocate for a more transparent approach, which we refer to as \textbf{Generation-time Fine-grained Provenance}. Unlike simple citation generation, this task requires the model to function as a transparent reasoner. For every generated sentence, the model must simultaneously identify the specific supporting source sentence and explicitly classify the evidence usage relation as \textit{Quotation}, \textit{Compression}, or \textit{Inference} (Figure~\ref{fig:intro}).

A major obstacle to this goal is the lack of suitable training data. Most existing benchmarks provide only coarse document-level attribution, are designed for post-hoc analysis, or lack structured supervision over evidence-use relations~\cite{gao-2023-enabling-llms-citation, zhu-etal-2025-trove}. To bridge this gap, we construct \textbf{ReFInE} (\textbf{Re}lation-aware \textbf{F}ine-grained \textbf{In}terpretability \& \textbf{E}vidence). Unlike previous heuristic-based datasets, ReFInE is built via a rigorous human-in-the-loop pipeline where LLM-assisted proposals undergo multi-stage expert verification. This ensures the dataset accurately captures complex evidence usage patterns, serving as a reliable testbed for transparent generation.

Building on ReFInE, we propose \textbf{GenProve}, a training framework tailored for this objective. We observe that standard SFT is insufficient; models often struggle to balance the fluidity of the answer with the strict structural constraints of provenance triples. GenProve overcomes this by integrating GRPO~\cite{guo2025deepseek} with a novel multi-dimensional reward modeling strategy. Specifically, we design a composite reward that goes beyond simple text quality. While strictly adhering to the structural constraints learned during SFT, our objective explicitly optimizes for \textit{content fidelity} and \textit{provenance correctness}, penalizing hallucinations where the cited evidence does not semantically support the generated claim. This holistic alignment forces the model to treat citation not as a stylistic decoration, but as an intrinsic reasoning constraint.

We evaluate GenProve against 14 strong LLMs. Results show that GenProve establishes a new state-of-the-art, significantly outperforming competitors in both answer quality and provenance accuracy. Crucially, our diagnostic analysis exposes a reasoning gap where models easily master exact \textit{Quotation}, yet they struggle significantly with \textit{Inference}. This suggests that the reliability of logical deductions remains a key challenge for future research.

Our contributions are summarized as follows:
\begin{itemize}
  \item We define Generation-time Fine-grained Provenance, shifting from coarse document-level citations to sentence-level attribution with explicit relation typing.
  \item We release \textbf{ReFInE}, the first expert-annotated QA dataset that provides dense, typed provenance supervision for multi-document generation, enabling rigorous training and evaluation of model interpretability.
  \item We propose \textbf{GenProve}, integrating SFT with GRPO alignment to master structured provenance generation. Experiments demonstrate that GenProve establishes a new state-of-the-art, while our analysis reveals the difficulty of verifying inference-based claims compared to direct quotation.
\end{itemize}

\begin{figure*}[t]
    \centering
    \includegraphics[width=\linewidth]{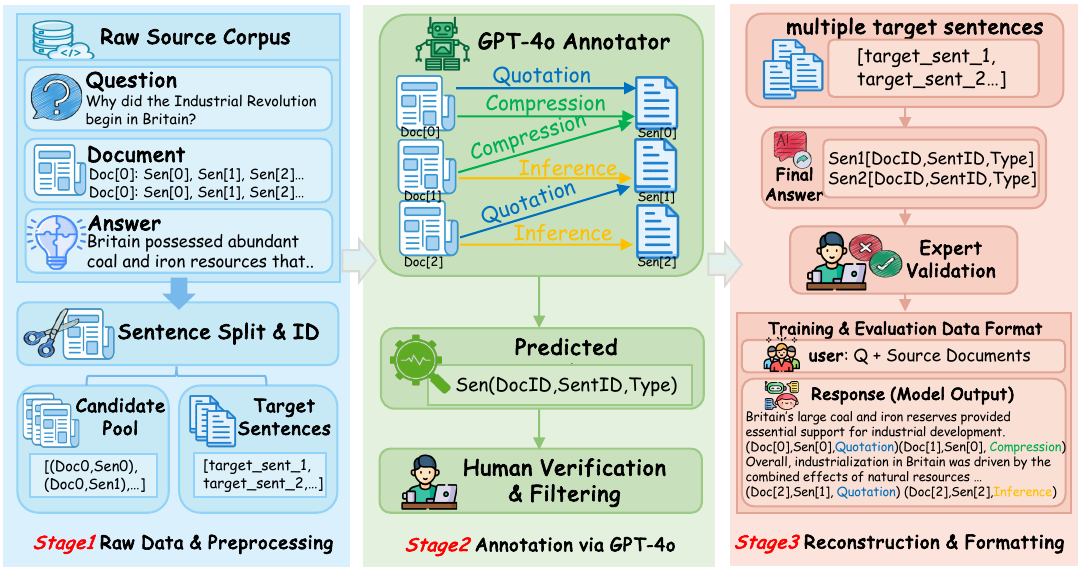}
    \caption{The construction pipeline of \textbf{ReFInE}. The process ensures high-quality provenance supervision through three stages: (1) preprocessing, (2) LLM-assisted annotation with filtering, and (3) reconstruction with rigorous \textbf{human-in-the-loop expert validation} to verify evidence sufficiency and relation correctness.}
    \label{fig:prove_asqa_pipeline}
\end{figure*}

\section{Related Work}

\textbf{Citation-Aware Text Generation.}
Incorporating citations into generated text is a critical step towards verifiable AI. Early approaches focus on stylistic imitation of academic writing~\cite{xing-2020-automatic-citation-texts} or utilize post-hoc retrieval to verify generated claims after the fact~\cite{li-2024-citation-enhanced,cao-2024-calm}.
With the advent of LLMs, the focus has shifted to RAG~\cite{gao-2023-enabling-llms-citation, wei2026mirage, li2026bayesrag}. Benchmarks like ALCE~\cite{gao-2023-enabling-llms-citation} have standardized the evaluation of citation quality, emphasizing document recall and precision. However, these methods typically operate at a coarse granularity, retrieving entire documents or passages without pinpointing the specific evidence used.
Recent training-based methods~\cite{aly-etal-2024-learning, slobodkin2024attribute} attempt to improve robustness by fine-tuning models to cite sources. However, these approaches are limited by treating citations as untyped pointers (e.g., simply linking to [1]). Our study advances this paradigm by enforcing typed relations, requiring the model to demonstrate an explicit understanding of the semantic relationship between the claim and the evidence, such as whether it is quoting or inferring.

\noindent\textbf{Fine-Grained Provenance \& Verification.} To improve interpretability, granularity in attribution has evolved from document-level to sentence-level. Previous works like GERE~\cite{chen2022gere} and SCIFI~\cite{cao2024verifiable} explore generating sentence identifiers, while \citet{kambhamettu2024traceable} introduce phrase-level links.
Most relevant to our work is TROVE~\cite{zhu-etal-2025-trove}, which introduces a comprehensive taxonomy for provenance relations. We \textbf{adopt their core taxonomy} (\textit{Quotation}, \textit{Compression}, \textit{Inference}) to ensure rigorous classification. However, we explicitly exclude their "Other" category. We omit this label for two reasons: first, it represents a negligible long-tail of the distribution; second, it serves as an ambiguous catch-all bucket. Such undefined signals lack clear semantic boundaries, making them unsuitable optimization targets for precise model alignment.
Furthermore, while TROVE focuses on post-hoc analysis of static text, GenProve targets \textbf{generation-time} provenance. We integrate these fine-grained types directly into the training process, shifting the paradigm from checking after generation to generating with inherent verification.

\noindent\textbf{Reasoning with Evidence.} Research studies investigate how LLMs reason with retrieved context. Studies like FRONT~\cite{huang2024learning} and SciRGC~\cite{li2025scirgc} have begun to model the latent reasoning process behind citations, inspired by chain-of-thought prompting.
GenProve pushes this direction further by explicitly supervising the \textit{Inference} relation. Unlike previous works that often conflate simple retrieval with complex reasoning, our framework distinguishes between surface-level copying and deep synthesis. By optimizing for specific relation types, we evaluate and improve the model's ability to abstract and deduce information rather than merely retrieving and copying verbatim segments.

\section{Dataset}

\subsection{Overview}

We study GenProve, a generation-time fine-grained provenance task where a system produces an answer together with sentence-level evidence links and typed provenance relations. A central obstacle to learning GenProve is the lack of suitable training data: existing benchmarks often provide only coarse document-level attribution, are designed for post-hoc analysis, or do not distinguish how evidence is used, such as quotation, compression, or inference. To address this gap, we construct ReFInE, a supervised dataset that provides multi-document inputs and reference answers annotated with structured provenance tags.

Each ReFInE instance pairs a user question with multiple source documents and a reference answer, where every sentence carries a provenance annotation linking it to source sentences via Quotation, Compression, or Inference. We build the dataset through a three-stage pipeline comprising sentence-level preprocessing, GPT-4o–based provenance annotation with human screening, and expert-validated reconstruction into a unified message format. Figure~\ref{fig:prove_asqa_pipeline} summarizes the construction process.

\subsection{Dataset construction}

\textbf{Raw data and sentence-level preprocessing}
We build ReFInE on top of a public long-form QA corpus with retrieved multi-document evidence~\cite{yehudai2024genie}, where each example contains a user question, a set of source documents, and a long-form reference answer (Figure~\ref{fig:prove_asqa_pipeline}, Stage~1). We choose this corpus because fine-grained typed provenance requires content-grounded, multi-document, long-form answers, while the original corpus does not provide sentence-level provenance or relation labels.
For each instance, we treat the user query as the question $Q$, segment the long answer into sentences to obtain a sequence of target sentences $\{t_1, t_2, \dots\}$, and segment all source documents into sentences. Each source sentence in $D$ is assigned a unique pair $(\text{Doc\_ID}, \text{Sent\_ID})$, which later serves as the indexing scheme for the provenance triples in Eq.~\ref{eq:provenance_set}. This stage produces a candidate pool of sentence-level evidence drawn from the source documents, together with a set of target sentences that require provenance labels.

\textbf{GPT-4o-based provenance annotation and preliminary filtering.} Given a question $Q$, a document set $D$, and a target sentence $t_j$, we prompt GPT-4o to predict provenance triples. These triples follow the (DocID, SentID, Relation) format, covering Quotation, Compression, and Inference (Figure~\ref{fig:prove_asqa_pipeline}, Stage 2). This process annotates each sentence independently to form a candidate pool. Next, three annotators screen the outputs for quality. They verify instruction compliance, fluency, and ethical safety. Crucially, they enforce strict [PROVE] formatting, checking for tag completeness, evidence merging, and index consistency. Violating samples are removed or corrected. Appendix~\ref{app:proveasqa-pre-filter} details this protocol.

\textbf{Reconstruction, expert validation, and final formatting.} We reconstruct instance-level examples by aggregating sentence-level annotations (Figure~\ref{fig:prove_asqa_pipeline}, Stage 3). Target sentences are sorted by their original order and concatenated, with each sentence receiving a [PROVE] tag that enumerates its evidence and relations (Eq.~\ref{eq:prove_example}). Subsequently, three experts conduct a second-round validation. Under a dual-check protocol, they rigorously verify whether each answer sentence is fully supported by its cited provenance set and whether each relation label (Quotation, Compression, Inference) is correct; instances failing either check are revised or discarded. Finally, valid samples are formatted for training: the user message comprises the question $Q$ and documents $D$, while the assistant message contains the long answer $A$ with embedded [PROVE] tags. Appendix~\ref{app:proveasqa-expert-validation} details this protocol.

\subsection{Dataset Analysis}
\label{subsec:proveasqa-analysis}

\begin{figure}[t]
  \centering
  \setlength{\abovecaptionskip}{0.12cm}    
  \setlength{\belowcaptionskip}{-0.6cm}
  \includegraphics[width=0.85\linewidth]{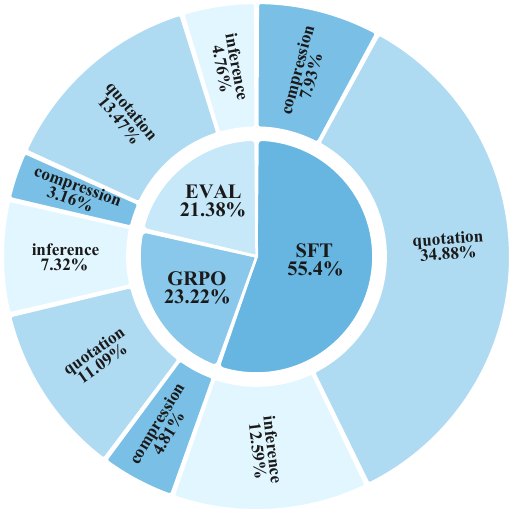}
  \caption{Relation-type distribution in ReFInE.}
  \label{fig:proveasqa_dist}
\end{figure}

\textbf{Split composition and relation distribution.}
ReFInE consists of three subsets that support SFT, RL-based training (GRPO), and held-out evaluation (EVAL), containing 12{,}540, 5{,}256, and 4{,}838 instances, respectively. Figure~\ref{fig:proveasqa_dist} summarizes the split composition and the relation-type mixture within each subset. The split proportions are 55.4\% (SFT), 23.22\% (GRPO), and 21.38\% (EVAL). The outer ring further shows that Quotation dominates across all splits, whereas GRPO allocates a larger share to Inference and Compression than EVAL, making it more suitable for optimizing reasoning and abstraction under sentence-level evidence constraints.

\textbf{Provenance density and corpus-level statistics.}
In ReFInE, each answer contains 3.96 provenance tags on average, with each tag aggregating 1.98 provenance triples. Full corpus-level statistics are reported in Appendix~\ref{app:dataset_stats}.

\subsection{Comparison with Existing Benchmarks}
\label{subsec:proveasqa_comparison}

Table~\ref{tab:proveasqa_benchmark_comparison} compares ReFInE with representative citation-aware and provenance benchmarks along axes central to fine-grained accountability, including relation expressivity, provenance granularity, task form, and whether provenance is produced alongside the answer using structured tags.

\textbf{Fine-grained, typed sentence-level provenance.}
Many prior benchmarks emphasize attribution but collapse provenance into a single untyped support signal or operate at a coarser granularity, which limits sentence-level inspection and weakens the interpretability of how evidence supports each claim.
In contrast, ReFInE annotates each answer sentence with sentence-level evidence links and explicit relation types, enabling relation-aware auditing beyond identifying the source alone.
\begin{figure*}[t]
  \centering
  \includegraphics[width=\linewidth]{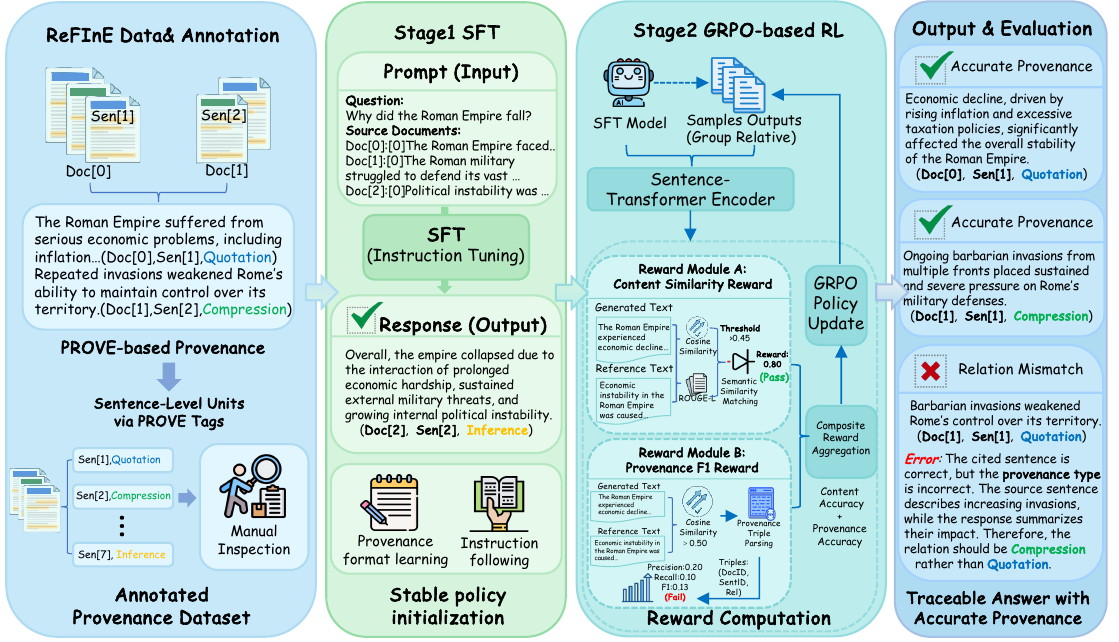}
  \caption{The \textbf{GenProve} framework. The model first undergoes SFT for instruction following and format learning. It is then aligned using GRPO with a \textbf{composite reward mechanism} that jointly optimizes for answer fidelity (content similarity reward) and fine-grained provenance accuracy (F1 Reward).}
  \label{fig:genprove_pipeline}
\end{figure*}

\textbf{Generation-time supervision.}
Several settings perform provenance analysis post hoc or append/verify citations in multi-step pipelines. ReFInE instead requires the answer and its provenance to be produced simultaneously, providing direct supervision and evaluation for generation-time provenance-aware decoding. Full benchmark comparison details are provided in Appendix~\ref{app:benchmark_comparison}.

\section{Method}
\label{sec:method}

We develop GenProve, a two-step training framework that enables a model to generate answers with fine-grained provenance. Given a question $Q$ and a document collection $D=\{d_1,\dots,d_m\}$, the model produces an answer $A=(t_1,\dots,t_n)$. Each answer sentence $t_j$ is accompanied by a set of provenance triples that identify supporting source sentences in $D$ and their relation types. Figure~\ref{fig:genprove_pipeline} shows the overall procedure.

\subsection{Supervised Fine-Tuning}
\label{subsec:sft}

We treat SFT as a foundational warm-up stage primarily designed to enforce structure adherence and provide stable format grounding. It trains the base model to follow instructions and to emit well-formed provenance annotations together with fluent answers, so that the subsequent RL stage can focus on reward-driven refinement rather than basic schema learning. Let $\mathcal{D}_{\text{SFT}}=\{(Q_i, D_i, A^{\mathrm{ref}}_i)\}_{i=1}^{N}$ denote the training set constructed from ReFInE, where $A^{\mathrm{ref}}_i$ is the reference answer annotated with sentence-level provenance. We fine-tune the model by maximizing the conditional likelihood of the reference output:
\begin{equation}
\small
\mathcal{L}_{\text{SFT}}(\theta)
=
-\sum_{i=1}^{N} \log p_{\theta}\!\left(A^{\mathrm{ref}}_i \mid Q_i, D_i\right).
\label{eq:sft_loss}
\end{equation}
This step provides a stable policy initialization that reliably produces syntactically valid provenance tags and on-topic content. Crucially, this structural foundation enables the subsequent RL stage to focus on refining the model's provenance accuracy rather than struggling with basic formatting errors.

\subsection{GRPO-based Reinforcement Learning}
\label{subsec:grpo}

Reinforcement learning improves provenance accuracy and reduces unsupported statements by optimizing a reward that evaluates both content and provenance. Starting from the SFT policy $\pi_{\theta}$, we sample a group of candidate answers for each input $(Q,D)$ and update the policy using GRPO. The objective maximizes the expected reward:
\begin{equation}
\small
\mathcal{J}(\theta)
=
\mathbb{E}_{(Q,D)\sim \mathcal{D}_{\text{GRPO}}}
\Big[
\mathbb{E}_{A \sim \pi_{\theta}(\cdot \mid Q,D)}
\big[R(A, A^{\mathrm{ref}})\big]
\Big].
\label{eq:rl_objective}
\end{equation}
The reward $R(A, A^{\mathrm{ref}})$ aggregates two components: a sentence-matching content reward and a reference-guided provenance F1 reward (Figure~\ref{fig:genprove_pipeline}).

\paragraph{Reward Design.} GenProve computes rewards at sentence granularity by parsing provenance tags and splitting both the generated answer and the reference into sentence units. Let $A=(t_1,\dots,t_n)$ and $A^{\mathrm{ref}}=(t^{\mathrm{ref}}_1,\dots,t^{\mathrm{ref}}_M)$. We represent both answers as sentence--provenance pairs:
\begin{equation}
\small
\label{eq:split_parse}
\begin{cases}
A = \{(t_j, P_j)\}_{j=1}^{n},\\
A^{\mathrm{ref}} = \{(t^{\mathrm{ref}}_k, P^{\mathrm{ref}}_k)\}_{k=1}^{M}.
\end{cases}
\end{equation}
Each $P_j$ (or $P^{\mathrm{ref}}_k$) is a set of triples of the form $(\mathrm{doc\_id}, \mathrm{sent\_id}, r)$ with
$r \in \{\mathrm{Quotation}, \mathrm{Compression}, \mathrm{Inference}\}$.

\paragraph{Reward A: Sentence-matching content similarity.}
This reward encourages semantic alignment with the reference while preserving sentence-level structure. For each generated sentence $t_j$, we find the best-matching reference sentence by cosine similarity between sentence embeddings produced by a Sentence-Transformer encoder:

\begin{equation}
k(j)=\arg\max_{k \in \{1,\dots,M\}} \cos\big(\phi(t_j), \phi(t^{\mathrm{ref}}_k)\big).
\label{eq:best_match_content}
\end{equation}
Here $\phi(\cdot)$ denotes the encoder. If the best cosine score is below a threshold $\tau_c$, the reward for this sentence is zero; otherwise we compute ROUGE-L between the matched pair:
\begin{equation}
\small
\begin{split}
r_{\text{sim}}(t_j) &=
\mathbb{I}\Big[\cos\big(\phi(t_j), \phi(t^{\mathrm{ref}}_{k(j)})\big) \ge \tau_c \Big] \\
&\quad \cdot \mathrm{ROUGE\text{-}L}\big(t_j, t^{\mathrm{ref}}_{k(j)}\big).
\end{split}
\label{eq:sim_reward_sentence}
\end{equation}
The content reward is the mean across sentences:
\begin{equation}
\small
R_{\text{sim}}(A,A^{\mathrm{ref}})
=
\frac{1}{n}\sum_{j=1}^{n} r_{\text{sim}}(t_j).
\label{eq:sim_reward}
\end{equation}

\paragraph{Reward B: Reference-guided provenance F1.}
This reward encourages generating correct provenance triples and relation types. To reduce missed provenance, we align sentences inversely. For each reference sentence $t^{\mathrm{ref}}_k$, we retrieve the closest generated sentence by cosine similarity:
\begin{equation}
j(k)=\arg\max_{j \in \{1,\dots,n\}}
\cos\big(\phi(t^{\mathrm{ref}}_k), \phi(t_j)\big).
\label{eq:best_match_cite}
\end{equation}
We gate mismatched pairs using a similarity threshold $\tau_p$. Given an aligned pair, we compare their provenance sets. Let
$I_k = P_{j(k)} \cap P^{\mathrm{ref}}_k$ denote the set of correctly reproduced provenance triples.
We compute sentence-level precision and recall as
$\mathrm{Prec}_k = |I_k| / |P_{j(k)}|$ and
$\mathrm{Rec}_k = |I_k| / |P^{\mathrm{ref}}_k|$,
and define the provenance score by
\begin{equation}
\small
F1_k =
\frac{2\,\mathrm{Prec}_k\,\mathrm{Rec}_k}
{\mathrm{Prec}_k+\mathrm{Rec}_k+\epsilon}.
\label{eq:cite_f1}
\end{equation}
The provenance reward averages sentence-level scores over all reference sentences, while gating out mismatched pairs:
\begin{equation}
\small
\label{eq:prov_reward}
\resizebox{0.9\hsize}{!}{$ 
\begin{aligned}
R_{\text{prov}}(A,A^{\mathrm{ref}})
=
\frac{1}{M}\sum_{k=1}^{M}
\mathbb{I}\!\left[
\cos\big(\phi(t^{\mathrm{ref}}_k), \phi(t_{j(k)})\big)
\ge \tau_p
\right]
\cdot F1_k .
\end{aligned}
$}
\end{equation}

\paragraph{Composite reward.}
We combine the two components into a single scalar reward:
\begin{equation}
\small
R(A,A^{\mathrm{ref}})
=
\alpha\,R_{\text{sim}}(A,A^{\mathrm{ref}})
+\beta\,R_{\text{prov}}(A,A^{\mathrm{ref}}),
\label{eq:composite_reward}
\end{equation}
where $\alpha$ and $\beta$ balance content fidelity and provenance correctness. This design penalizes common failure modes shown in Figure~\ref{fig:genprove_pipeline}, including incorrect relation typing and unsupported or out-of-document provenance.

\section{Experiments}

\subsection{Experimental Setup}
\label{subsec:exp_setup}

\noindent\textbf{Models.}
We evaluate 14 LLMs, covering both open- and closed-source systems. The open-source models include Llama-3.1-8B-Instruct~\cite{dubey2024llama}, Gemma-3-12B-it~\cite{team2025gemma}, Yi-1.5-9B-Chat~\cite{liu2025open}, Qwen3-8B~\cite{yang2025qwen3}, InternLM2.5-7B-Chat~\cite{cai2024internlm2}, Hunyuan-7B-Instruct~\cite{zheng2025hunyuan}, Vicuna-7B-v1.5~\cite{zheng2023judging}, Baichuan2-7B-Chat~\cite{yang2023baichuan}, Qwen3-14B~\cite{yang2025qwen3}, GLM-4-9B~\cite{glm2024chatglm}, and GLM-4.5~\cite{glm2024chatglm}. The closed-source models include Gemini~2.5~Pro~\cite{comanici2025gemini}, GPT-5~\cite{achiam2023gpt}, and Kimi~\cite{team2025kimi}. All models use a unified input format of questions and source documents. 
The full inference prompt is given in Appendix~\ref{subsec:appendix_inference_prompt}.

\textbf{Training configuration.}
GenProve is trained in two steps. For supervised fine-tuning, we start from Qwen3-8B~\cite{yang2025qwen3} and perform full-parameter optimization with AdamW, using a learning rate of $2\times10^{-5}$, a maximum sequence length of 2048, and gradient accumulation to achieve an effective batch size of 16.
For GRPO alignment, we initialize from the SFT model and continue optimization under the same learning rate and sequence length settings, with temperature set to 1, $\beta=0.02$, and 4 iterations per update.
For reward computation, the sentence-matching and provenance-alignment thresholds are set to $\tau_c=0.45$ and $\tau_p=0.50$, respectively.

\begin{table*}[t]
\centering
    \resizebox{\textwidth}{!}{
    \begin{tabular}{l|cccc|ccc|c|c}
    \toprule
    Model & ROUGE-L$\uparrow$ & BLEU$\uparrow$ & METEOR$\uparrow$ & MoverScore$\uparrow$ & Prec.$\uparrow$ & Rec.$\uparrow$ & F1$\uparrow$ & Format (\%)$\uparrow$ & LLM-as-judge (1--5)$\uparrow$ \\
    \midrule
    \midrule
    Baichuan2-7B~\cite{yang2023baichuan} & 34.68  & 19.41  & 48.08  & 32.03  & 3.22  & 3.41  & 3.04  & 26.60  & 0.78  \\
    Vicuna-7b-v1.5~\cite{zheng2023judging} & 38.83  & 24.37  & 50.08  & 34.73  & 9.01  & 5.85  & 6.70  & 92.40  & 1.10  \\
    InternLM2.5-7B~\cite{cai2024internlm2} & 47.79  & 30.60  & 53.47  & 43.60  & 12.64  & 13.35  & 11.94  & 80.24  & 1.67  \\
    Hunyuan-7B~\cite{zheng2025hunyuan} & 39.91  & 24.70  & 43.74  & 32.99  & 22.78  & 22.86  & 21.43  & 90.88  & 1.72  \\
    Yi-1.5-9B~\cite{liu2025open} & 48.26  & 30.11  & 47.77  & 43.35  & 20.71  & 22.91  & 20.11  & 96.81  & 1.80  \\
    Llama-3.1-8B~\cite{dubey2024llama} & 47.71  & 26.36  & 42.04  & 41.14  & 21.78  & 20.30  & 20.05  & \cellcolor[rgb]{ .933,  .973,  .91}99.85  & 2.00  \\
    GLM-4-9B~\cite{glm2024chatglm} & 50.33  & 32.62  & 50.00  & 44.93  & 34.57  & 34.12  & 32.79  & \cellcolor[rgb]{ .992,  .941,  .902}\textbf{100.0 } & 2.16  \\
    Qwen3-8B~\cite{yang2025qwen3} & 51.80  & 35.56  & 55.38  & 45.90  & 37.78  & 30.56  & 32.53  & \cellcolor[rgb]{ .992,  .941,  .902}\textbf{100.0 } & 2.25  \\
    Gemma-3-12B~\cite{team2025gemma} & 48.97  & 32.13  & 50.80  & 43.79  & 41.06  & 31.11  & 34.03  & \cellcolor[rgb]{ .992,  .941,  .902}\textbf{100.0 } & 2.47  \\
    Qwen3-14B~\cite{yang2025qwen3} & \cellcolor[rgb]{ .933,  .973,  .91}52.85  & \cellcolor[rgb]{ .933,  .973,  .91}35.70  & 55.34  & \cellcolor[rgb]{ .933,  .973,  .91}47.06  & 45.80  & 40.33  & 41.16  & 99.70  & 2.59  \\
    GLM-4.5-355B~\cite{glm2024chatglm} & 49.81  & 35.05  & \cellcolor[rgb]{ .933,  .973,  .91}57.69  & 44.92  & \cellcolor[rgb]{ .933,  .973,  .91}48.84  & \cellcolor[rgb]{ .933,  .973,  .91}44.03  & \cellcolor[rgb]{ .933,  .973,  .91}44.55  & 98.63  & \cellcolor[rgb]{ .933,  .973,  .91}2.63  \\
    \midrule
    \midrule
    Kimi~\cite{team2025kimi}  & 49.33  & 31.24  & 51.56  & 43.84  & 31.55  & 32.09  & 29.70  & 99.09  & 2.20  \\
    GPT-5~\cite{achiam2023gpt} & 41.79  & 20.01  & 38.83  & 36.38  & 21.37  & 16.73  & 17.88  & \cellcolor[rgb]{ .933,  .973,  .91}99.85  & 2.23  \\
    Gemini 2.5 Pro~\cite{comanici2025gemini} & 48.75  & 31.77  & 53.09  & 44.79  & 46.68  & 42.86  & 42.92  & \cellcolor[rgb]{ .992,  .941,  .902}\textbf{100.0 } & 2.57  \\
    \midrule
    \midrule
    GenProve (Ours) & \cellcolor[rgb]{ .992,  .941,  .902}\textbf{57.25 } & \cellcolor[rgb]{ .992,  .941,  .902}\textbf{42.22 } & \cellcolor[rgb]{ .992,  .941,  .902}\textbf{59.39 } & \cellcolor[rgb]{ .992,  .941,  .902}\textbf{51.04 } & \cellcolor[rgb]{ .992,  .941,  .902}\textbf{54.96 } & \cellcolor[rgb]{ .992,  .941,  .902}\textbf{51.26 } & \cellcolor[rgb]{ .992,  .941,  .902}\textbf{51.21 } & \cellcolor[rgb]{ .933,  .973,  .91}99.85  & \cellcolor[rgb]{ .992,  .941,  .902}\textbf{3.14 } \\
    \bottomrule
    \end{tabular}%
    }
  \caption{Main results on ReFInE. Our proposed GenProve consistently outperforms strong open-source and closed-source LLMs across answer quality, provenance accuracy, and LLM-based evaluation.}
  \label{tab:main_results}%
\end{table*}%

\textbf{Evaluation Metrics.} We evaluate models along three axes: answer quality, provenance accuracy, and format validity. Answer quality is measured using ROUGE-L~\cite{lin-2004-rouge}, BLEU~\cite{papineni-etal-2002-bleu}, METEOR~\cite{banerjee-lavie-2005-meteor}, BERTScore~\cite{Zhang2020BERTScore}, and MoverScore~\cite{zhao-etal-2019-moverscore}, computed on answers excluding provenance tags. Provenance accuracy is assessed by sentence-level precision, recall, and F1 via exact matching over document id, sentence id, and relation type, while format validity reports the percentage of outputs that strictly follow the required provenance schema. Additionally, we conduct subjective evaluations with LLM and human judges: the former provides relation-specific scores, while the latter assesses answer quality and provenance correctness (prompts and guidelines in Appendices~\ref{subsec:appendix_llm_judge}--\ref{subsec:appendix_human_eval}).

\subsection{Main Results}
\label{subsec:main_results}
Table~\ref{tab:main_results} presents the main results on ReFInE. GenProve achieves the best overall performance and ranks first on all evaluation axes, including answer quality, provenance accuracy, and the LLM-as-judge score. Notably, it also achieves the highest provenance recall, indicating stronger evidence coverage in addition to higher precision and F1. The gains are consistent across automatic metrics and subjective judging, demonstrating that generation-time fine-grained provenance training improves both the usefulness of answers and the reliability of sentence-level provenance.

Across model groups, open-source systems exhibit substantial variance. Earlier chat-style or lightly instruction-tuned models, such as Baichuan2-7B and Vicuna-7B-v1.5, often fail to follow the provenance schema, leading to low format validity and weak provenance accuracy. In contrast, more recent open-source models, including Qwen3 and GLM-4, generate valid outputs more consistently and achieve markedly higher provenance F1 and LLM-judge scores. Among non-GenProve systems, GLM-4.5 is the strongest baseline, ranking second in both provenance quality and LLM-judge score. Closed-source models are competitive: Gemini~2.5~Pro is the strongest closed-source baseline, but still trails GenProve on the overall judge score. This pattern is especially notable for some closed-source models, whose strong fluency does not always translate into strong fine-grained provenance, a point we analyze further in Section~\ref{subsec:diagnostic}.

From the metric perspective, answer-quality metrics show that GenProve generates more faithful and fluent responses after provenance tags are removed. It exceeds the strongest baseline on ROUGE-L, BLEU, METEOR, and MoverScore, indicating improvements in both surface overlap and semantic similarity. Provenance metrics show the largest margin: GenProve achieves a substantially higher provenance F1 than the strongest baseline, suggesting more accurate sentence-level evidence localization and relation typing. Because provenance scores are computed by averaging over reference sentences, these gains reflect broader and more accurate evidence coverage rather than redundant repetition of a few supported claims. Correct format highlights that formatting is necessary but not sufficient: several strong baselines already achieve near-perfect parseability, whereas weaker baselines fail frequently; GenProve maintains similarly high compliance. Finally, the LLM-as-judge score summarizes end-to-end quality under joint requirements of correctness, fluency, and traceability, where GenProve attains the highest overall score.

\begin{table}[t]
\centering
\small
\resizebox{\linewidth}{!}{
\begin{tabular}{lccc}
\toprule
Model & Doc-Precision$\uparrow$ & Doc-Recall$\uparrow$ & Doc-F1$\uparrow$ \\
\midrule
VANILLA~\cite{gao-2023-enabling-llms-citation} & 72.50 & 73.60 & 73.05 \\
SUMM~\cite{gao-2023-enabling-llms-citation} & 68.90 & 61.80 & 65.16 \\
SNIPPET~\cite{gao-2023-enabling-llms-citation} & 65.30 & 57.40 & 61.10 \\
APO~\cite{li2024improving} & 69.60 & 72.80 & 71.20 \\
VeriCite~\cite{qian2025vericite} & 74.61 & 81.13 & 77.73 \\
GERE~\cite{chen2022gere}  & 84.43 & 78.01 & 81.10 \\
\midrule
\textbf{GenProve (Ours)} & \textbf{94.45} & \textbf{93.16} & \textbf{92.43} \\
\bottomrule
\end{tabular}
}
\caption{Document-level citation benchmark. We collapse sentence-level provenance to document IDs only and compare against document-level citation systems.}
\label{tab:doc_level_benchmark}
\end{table}

\begin{table}[t]
\small
\setlength{\belowcaptionskip}{-0.3cm}
  \centering
  \resizebox{\linewidth}{!}{
  \begin{tabular}{lccccc}
    \toprule
    \textbf{Model} & \textbf{BLEU} & \textbf{BERTScore} & \textbf{F1} & \textbf{Format} & \textbf{Judge} \\
    \midrule
    GenProve & 42.22 & 61.98 & 51.21 & 99.85 & 3.14 \\
    w/o Prov Reward & 11.94 & 46.96 & 24.67 & 96.66 & 2.20 \\
    w/o Sim Reward & 24.60 & 44.50 & 60.32 & 95.74 & 2.71 \\
    w/o GRPO & 41.82 & 60.70 & 50.48 & 99.70 & 2.62 \\
    \bottomrule
  \end{tabular}
  }
  \caption{Ablation study on ReFInE. The results validate the necessity of GRPO alignment and the complementary roles of content similarity and provenance rewards.}
  \label{tab:ablation_main}
\end{table}

\begin{figure*}[t]
  \centering
  \includegraphics[width=\linewidth]{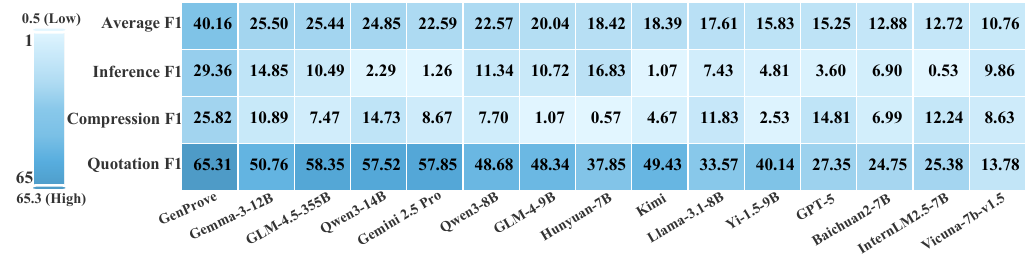}
  \caption{Performance breakdown by relation type (F1 score). The heatmap reveals a reasoning gap: while most models handle verbatim \textit{Quotation} well, they struggle significantly with \textit{Inference}. \textbf{GenProve} consistently outperforms baselines, showing the largest gains in complex provenance tasks (Compression and Inference).}
  \label{fig:relation_type_f1}
\end{figure*}

\subsection{Document-Level Citation Benchmark}
\label{subsec:doc_level_benchmark}

For direct comparison with prior document-level citation methods, we additionally evaluate GenProve under a DocID-only setting by collapsing sentence-level provenance to cited document sets.
Table~\ref{tab:doc_level_benchmark} shows that GenProve achieves the best performance on all three metrics, reaching 94.45 Doc-Precision, 93.16 Doc-Recall, and 92.43 Doc-F1.
Compared with the previous state of the art, GERE, this yields a gain of 11.33 Doc-F1.
Although GenProve is trained for sentence-level typed provenance, its strong performance under this coarse-grained setting suggests robust cross-granularity transfer.

\subsection{Ablation Study}
\label{subsec:ablation}

\begin{figure}[t]
  \centering
  \setlength{\belowcaptionskip}{-0.3cm}
  \includegraphics[width=\linewidth]{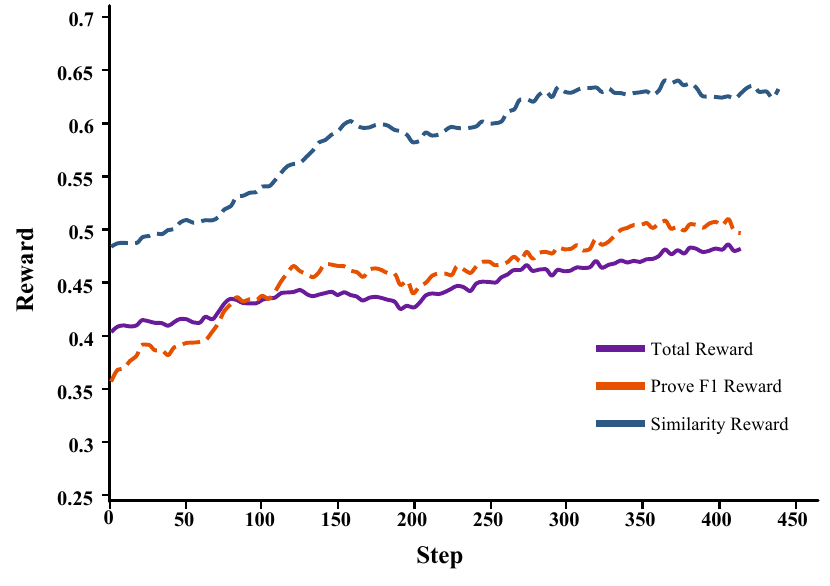}
  \caption{Learning dynamics during GRPO. Consistent upward trends in content similarity and provenance F1 rewards indicate GenProve improves provenance reliability without compromising answer faithfulness, achieving coordinated optimization of dual objectives.}
  \label{fig:grpo_dynamics}
\end{figure}
\begin{figure}[t]
  \centering
  \setlength{\abovecaptionskip}{0.12cm}    
  \setlength{\belowcaptionskip}{-0.6cm}
  \includegraphics[width=\linewidth]{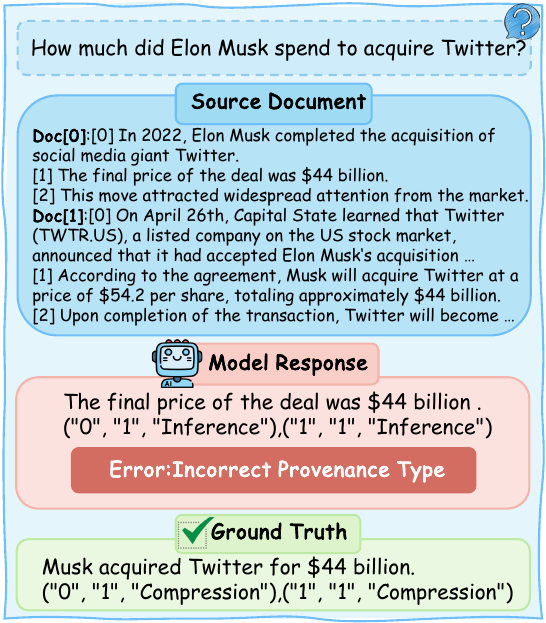}
  \caption{A representative relation-type error. The cited evidence is relevant, but the predicted provenance type does not match how the answer sentence is actually supported by the source.}
  \label{fig:error_type_main}
\end{figure}
Table~\ref{tab:ablation_main} shows that GRPO alignment significantly boosts end-to-end quality, raising judge scores substantially over the SFT baseline. Reward ablations further confirm the two components are complementary. Removing the provenance reward drops provenance F1 and judge scores, implying similarity alone cannot enforce precise provenance. Conversely, removing the similarity reward improves F1 yet harms answer quality, showing provenance optimization alone is insufficient. Combining both maximizes the judge score, effectively balancing fluent answers with correct, typed provenance.

\subsection{Diagnostic Analysis}
\label{subsec:diagnostic}

\textbf{Performance by Relation Type.}
Figure~\ref{fig:relation_type_f1} reports F1 by provenance relation type, reflecting the reliability of sentence-level provenance beyond verbatim reuse. The heatmap reveals a clear difficulty ordering: Quotation is easiest, while Compression and Inference are substantially harder, indicating challenges in evidence abstraction and integration. GenProve achieves the strongest performance across all three relations and the highest average F1, with its largest gains on Compression and Inference, reflecting improved evidence localization and relation typing in harder cases.

\noindent\textbf{GRPO Training Dynamics.}
Figure~\ref{fig:grpo_dynamics} visualizes reward trajectories during GRPO alignment. Both component rewards increase and stabilize, indicating the policy improves content alignment and provenance correctness jointly rather than oscillating between objectives. The total reward follows this upward trend, mirroring the complementary roles found in ablations: similarity optimization strengthens faithfulness, whereas PROVE-F1 strengthens typed provenance, and their combination supports superior overall quality.

\noindent\textbf{Qualitative Error Analysis.}
Figure~\ref{fig:error_type_main} presents a representative relation-type error in fine-grained provenance generation.
In this case, the model identifies relevant supporting evidence but assigns an incorrect provenance type, confusing \textit{Quotation}, \textit{Compression}, and \textit{Inference}.
This example shows that accurate provenance generation requires more than locating relevant evidence: the model must also correctly determine how the cited source supports the claim.
Such errors are more likely to arise when the support relation involves abstraction or reasoning rather than direct lexical overlap, which further reflects the difficulty of typed provenance generation.

\subsection{Consistency with Human Evaluation}
\label{subsec:human_consistency}
\begin{figure}[t]
  \centering
  \setlength{\belowcaptionskip}{-0.4cm}
  \includegraphics[width=\linewidth]{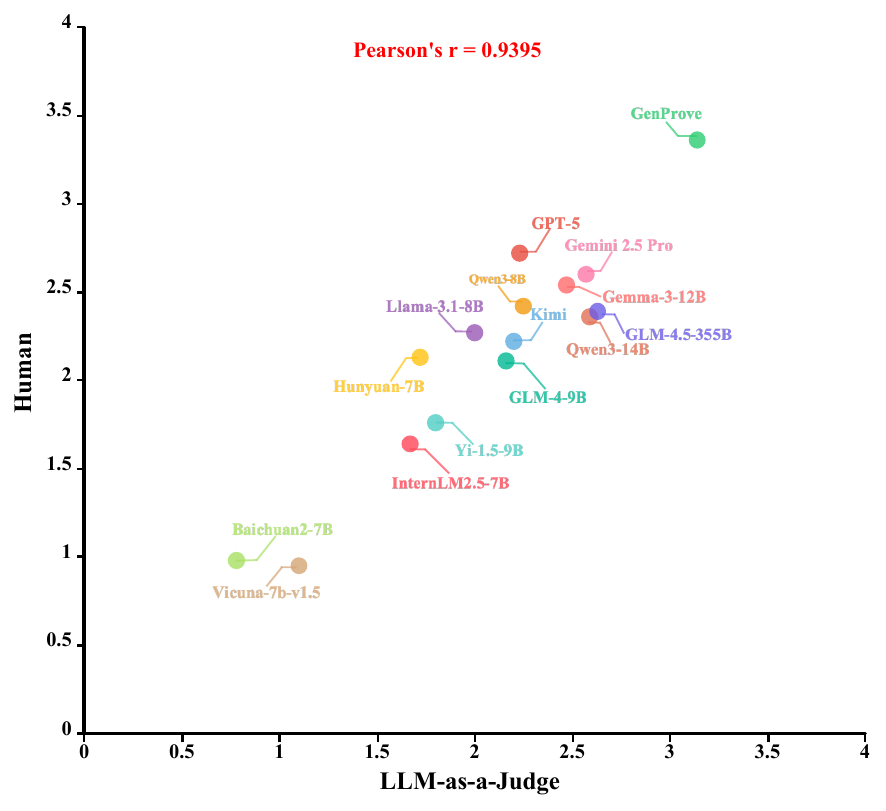}
  \caption{Correlation between LLM-as-a-Judge scores and human ratings. The high Pearson correlation ($r=0.9395$) validates our automatic metric. Notably, \textbf{GenProve} occupies the top-right corner, demonstrating significantly superior performance over all baselines under both automated and human evaluations.}
  \label{fig:judge_human_corr}
\end{figure}

To assess the reliability of LLM-as-a-Judge as our primary traceability-oriented evaluation signal, we measure its consistency with human evaluation at the model level.
Figure~\ref{fig:judge_human_corr} shows a strong positive correlation between LLM-as-a-Judge scores and human ratings, with a Pearson correlation coefficient of $r=0.9395$. 
This result indicates that the automatic judge closely aligns with human preferences under the same provenance-aware evaluation criteria, supporting its use for large-scale comparison in the main experiments. 
Detailed human evaluation results and scoring analyses are provided in Appendix~\ref{subsec:appendix_llm_judge_table} and ~\ref{subsec:appendix_human_results}.

\section{Conclusion}
We introduce a paradigm shift from coarse citations to generation-time fine-grained provenance. By constructing the \textbf{ReFInE} dataset and developing the \textbf{GenProve} framework, we demonstrate that LLMs can be trained to transparently document their evidence usage via structured triples. Experiments confirm that GenProve balances generation quality with strict provenance constraints, establishing a new state-of-the-art across 14 strong LLMs. Despite these advances, the performance gap between simple \textit{Quotation} and complex \textit{Inference} suggests that verifiable reasoning remains a frontier challenge. We position ReFInE as a stepping stone towards self-auditing LLMs, models that not only generate knowledge but explicitly reason about the provenance of their own assertions.

\section*{Limitations}
While GenProve establishes a new standard for fine-grained provenance, we identify three limitations to address in future work.
(1) \textbf{Inference latency}. Generating structured provenance triples inevitably increases the output token count compared to standard generation. Although essential for trustworthiness, this introduces a slight latency trade-off in real-time applications. (2) \textbf{Linguistic scope}. Our current ReFInE dataset and evaluation primarily focus on English. Extending the \textit{Quotation-Compression-Inference} taxonomy to multilingual or cross-lingual settings remains an open avenue for research.
(3) \textbf{Retrieval dependency}. Our framework focuses on the generation stage. Like all RAG systems, end-to-end performance is bounded by the quality of the retriever; if retrieved documents contain no relevant information, the model cannot generate valid provenance.

\section*{Acknowledgements}
We sincerely thank the anonymous reviewers, the area Chairs, and the program committee for their insightful comments, constructive suggestions, and efforts in organizing the review process.

\bibliography{custom}

\clearpage
\appendix

\section{Task Definition}

We study a text provenance task in which a system answers a user question and, for each answer sentence, specifies which source sentences support it and how they relate. Formally, given a question $Q$ and a collection of source documents $D = \{d_1, d_2, \dots, d_m\}$, where each document $d_i$ is a sequence of sentences $d_i = \{s_{i,1}, s_{i,2}, \dots, s_{i,k_i}\}$, the system produces an answer $A = (t_1, t_2, \dots, t_n)$ and a provenance set $P = (P_1, P_2, \dots, P_n)$ aligned with the answer. For each answer sentence $t_j$, the provenance $P_j$ is a set of triples:
\begin{equation}
P_j = \{(\text{doc\_id}, \text{sent\_id}, r)\},
\label{eq:provenance_set}
\end{equation}
where $\text{doc\_id}$ indexes a document in $D$, $\text{sent\_id}$ indexes a sentence within that document, so that $(\text{doc\_id}, \text{sent\_id})$ uniquely identifies a source sentence $s_{i,\ell}$ in $D$, and $r$ denotes the provenance relation type.

Each answer sentence $t_j$ may be supported by multiple source sentences, possibly drawn from different documents, and different links may carry different relation types. We consider three provenance relation types between an answer sentence $t_j$ and a source sentence $s_{i,\ell}$: \textbf{Quotation}, where $t_j$ copies or closely paraphrases the wording of $s_{i,\ell}$; \textbf{Compression}, where $t_j$ summarizes or paraphrases information that is distributed across one or more source sentences, such as $s_{i,\ell_1}$ and $s_{i,\ell_2}$; and \textbf{Inference}, where $t_j$ states a conclusion that is logically supported by one or more source sentences, such as $s_{i,\ell}$. For a single answer sentence, different supporting source sentences may be associated with different relation types, and all corresponding triples are collected in $P_j$ as in Eq.~\ref{eq:provenance_set}.

In our setting, the answer and its provenance are presented in a structured textual format by interleaving each answer sentence $t_j$ with a provenance annotation that enumerates the triples in $P_j$. Concretely, an answer sentence and its provenance may be rendered as:
\begin{equation}
\begin{split}
{\small Sentence. [PROVE: (d1, s6, 'Quotation'),} \\
{\small \qquad (d2, s3, 'Inference')]}
\end{split}
\label{eq:prove_example}
\end{equation}
where Sentence. corresponds to an answer sentence $t_j$, each pair (d1, s6) or (d2, s3) is a concrete instance of $(\text{doc\_id}, \text{sent\_id})$ in Eq.~\ref{eq:provenance_set}, and the strings 'Quotation' and 'Inference' instantiate the relation type $r$. The task thus requires a system to generate an answer to $Q$ and, at the same time, provide fine-grained, sentence-level provenance that identifies supporting source sentences and labels the type of relationship for every answer sentence.

\section{Additional Details of ReFInE Construction}
\label{app:proveasqa-construction}

\subsection{Preliminary filtering criteria}
\label{app:proveasqa-pre-filter}

After GPT-4o produces sentence-level answers and provenance candidates, all instances undergo a preliminary filtering stage conducted by three annotators. The goal is to remove clearly unusable or structurally invalid samples before expert validation. We adopt four main criteria.

\paragraph{Instruction compliance.}
The generated answer must directly address the user question. Annotators discard responses that ignore the query, explicitly refuse to answer (e.g., ``I do not know''), or simply restate the question without providing new information.

\paragraph{Fluency and completeness.}
Annotators remove outputs that exhibit severe grammatical errors, broken sentence structure, heavy repetition, or truncated content due to length limits that render the answer semantically incomplete.

\paragraph{Ethical compliance.}
Any sample containing hate speech, discriminatory language, personal identifiable information, or harmful recommendations is removed to ensure that ReFInE does not propagate unsafe content.

\paragraph{Format validity.}
We enforce strict constraints on the structure of provenance annotations. Annotators check:
\begin{itemize}
    \item \textbf{Tag completeness}: every factual sentence in the answer must be accompanied by a parsable [PROVE] tag; samples with missing or unparsable tags are discarded.
    \item \textbf{Merged multi-source references}: if a single sentence is supported by multiple evidence sentences, all evidence must appear within a single [PROVE:~(...)] block. Instances that split evidence across multiple tags for the same sentence are considered format violations.
    \item \textbf{Index consistency}: all DocID and SentID values must follow the zero-based indexing scheme and stay within the valid ranges of the input documents and sentences; out-of-bounds or misaligned indices lead to removal.
    \item \textbf{Tuple well-formedness}: each provenance tuple must contain exactly three fields ("doc\_id", "sent\_id", "relation") with correct types; malformed tuples are grounds for discarding the sample.
\end{itemize}

\begin{figure*}[t]
\centering
\setlength{\abovecaptionskip}{0.05cm}    
\setlength{\belowcaptionskip}{-0.5cm}
\begin{promptbox}[Dataset Construction Prompt]

\scriptsize
You need to complete three fields in the dataset: \texttt{ground\_truth\_global}, \texttt{ground\_truth\_local}, and \texttt{Candidate Text}. The specific tasks are:

\textbf{1. Analyze the relationship between the target sentence and each candidate sentence}  
Relationship types:  
• \textbf{Quotation}: The target sentence partially or fully replicates content from a candidate sentence, including exact quotes, slight edits, or incorporation of phrases.  
• \textbf{Compression}: The target sentence condenses information from one or more candidate sentences.  
• \textbf{Inference}: The target sentence is based on information implied rather than explicitly stated.  

\textbf{2. Populate the fields:}

\textbf{• ground\_truth\_global}:  
Key format: ``DocID-SentID'' → Relationship  
Only include candidate sentences relevant to the target sentence.

\textbf{• ground\_truth\_local}:  
Key format: local candidate index (based on \texttt{global2local\_id}) → Relationship  
Only include relevant candidate sentences.

\textbf{• Candidate Text}:  
A list of dictionaries:  
\{\texttt{"Text Address": "NULL", "Doc\_ID": "X", "Original Sentence": [\{\texttt{"Critical Sentence": "...", "Relationship": "...", "Sent\_ID": "..."}\}]}\}  
Include only sentences that support the target sentence.

---

\textbf{Example Input Object:}

\begin{verbatim}
{"id": -5742330000000000000,
 "target_id": 0,
 "target_sent": "\"Bunk'd\" was renewed for a third season by Disney Channel
 on August 31, 2017, and it premiered on June 18, 2018.",
 "Candidate Relationship Sets": ["Quotation", "Compression", "Inference"],
 "prompt_local": "[Content]=\"\"\"
Target Sentence: \"Bunk'd\" was renewed for a third season by Disney Channel on August 31, 2017, and it premiered on June 18, 2018.
Candidate Sentence [1]: \"The series was renewed for a third season by Disney Channel on August 31, 2017.
Candidate Sentence [2]: On June 1, 2018, it was announced that Peyton List, ...
Candidate Sentence [3]: The third season premiered on Disney Channel on June 18, 2018.
Candidate Sentence [4]: In March 2018, actress Skai Jackson stated ...
Candidate Sentence [5]: ...
\"\"\"
",
 "prompt_global": "[Content]=\"\"\"
Target Sentence: \"Bunk'd\" was renewed for a third season ...
Candidate Sentence [0-0]: \"The series was renewed for a third season ...
Candidate Sentence [0-1]: ...
Candidate Sentence [0-2]: The third season premiered on Disney Channel ...
Candidate Sentence [0-3]: ...
Candidate Sentence [1-0]: ...
Candidate Sentence [2-6]: The second season premiered on August 23, 2016.
\"\"\"
",
 "global2local_id": {"0-0": "1", "0-1": "2", "0-2": "3", ...},
 "ground_truth_global": {"0-0": "Quotation", "0-2": "Quotation"},
 "ground_truth_local": {"1": "Quotation", "3": "Quotation"},
 "Candidate Text": [
   {"Text Address": "NULL",
    "Doc_ID": "0",
    "Original Sentence": [
       {"Critical Sentence": "\"The series was renewed ... 2017.\"",
        "Relationship": "Quotation",
        "Sent_ID": "0"},
       {"Critical Sentence": "The third season premiered ... 2018.",
        "Relationship": "Quotation",
        "Sent_ID": "2"}
    ]}
 ]}
\end{verbatim}

\end{promptbox}
\caption{The prompt used in Stage 2 of dataset construction. It defines the schema for generating candidate provenance triples and mapping global-local indices for GPT-4o annotation.}
\label{fig:proveasqa-prompt}
\end{figure*}

\subsection{Expert validation protocol}
\label{app:proveasqa-expert-validation}

Instances that pass the preliminary filter then enter an expert validation stage. Three annotators with experience in NLP and factuality assessment independently review the remaining samples. For each answer sentence and its [PROVE] tag, annotators perform a two-part check that jointly considers evidence sufficiency and relation correctness.

\paragraph{Evidence sufficiency.}
Annotators examine whether the cited (DocID, SentID) tuples provide adequate support for the generated sentence. A provenance set is accepted if the content of the sentence can be fully derived from the cited source sentences without introducing unsupported external facts or contradicting the documents. If the answer contradicts the sources, lacks supporting evidence, or relies on hallucinated information, the instance is marked invalid and removed.

\paragraph{Relation correctness.}
Annotators standardize the use of the three relation types in ReFInE and verify that each predicted label matches the underlying evidence–sentence relationship.

\emph{Quotation.}
A link is labeled as Quotation when the answer sentence partially or fully copies the wording of the source sentence, allowing only minor grammatical adjustments such as tense or word order changes. If the sentence substantially rephrases or loosely paraphrases the source while being labeled as quotation, the label is corrected.

\emph{Compression.}
A link is labeled as Compression when the answer sentence is a faithful condensation of long or multi-sentence content in the source. Annotators check that the compressed sentence preserves the key information while shortening or simplifying the original wording. If a sentence merely copies the source or omits crucial information while being tagged as compression, the label is revised. For example, the sentence ``The dam releases water because of heavy rain'' can be accepted as a compression of a longer source description that explains rising water levels and forced discharge due to continuous rain.

\emph{Inference.}
A link is labeled as Inference when the answer states a conclusion that is logically supported by one or more source sentences. Annotators verify that the conclusion follows from the cited evidence, possibly requiring multi-hop reasoning, cross-paragraph or cross-document integration, or reasonable commonsense inference. If the answer introduces content that is not implied by the sources or breaks the reasoning chain, the instance is marked invalid. Typical accepted cases include multi-hop reasoning that combines two or more source sentences, cross-document aggregation of facts, and commonsense conclusions that extend but do not contradict the given evidence.

\subsection{Prompt for GPT-4o Annotation}
\label{app:proveasqa-gpt4o-prompt}

To ensure consistent sentence-level provenance annotation across ReFInE, we provide GPT-4o with a structured prompt that defines the three relation types, specifies the expected formats for ground\_truth\_global, ground\_truth\_local, and Candidate Text, and illustrates the mapping between global and local sentence indices. The prompt also includes an example input object that clarifies how candidate sentences and their relationships should be encoded. This prompt governs Stage~2 of the dataset construction pipeline, and the full prompt specification is shown in Figure~\ref{fig:proveasqa-prompt}, which standardizes all annotations prior to human verification.

\subsection{Examples of ReFInE Instances}
\label{app:proveasqa-examples}
\begin{figure}[t]
\setlength{\abovecaptionskip}{0.05cm}    
\setlength{\belowcaptionskip}{-0.5cm}
    \centering
    \includegraphics[width=\linewidth]{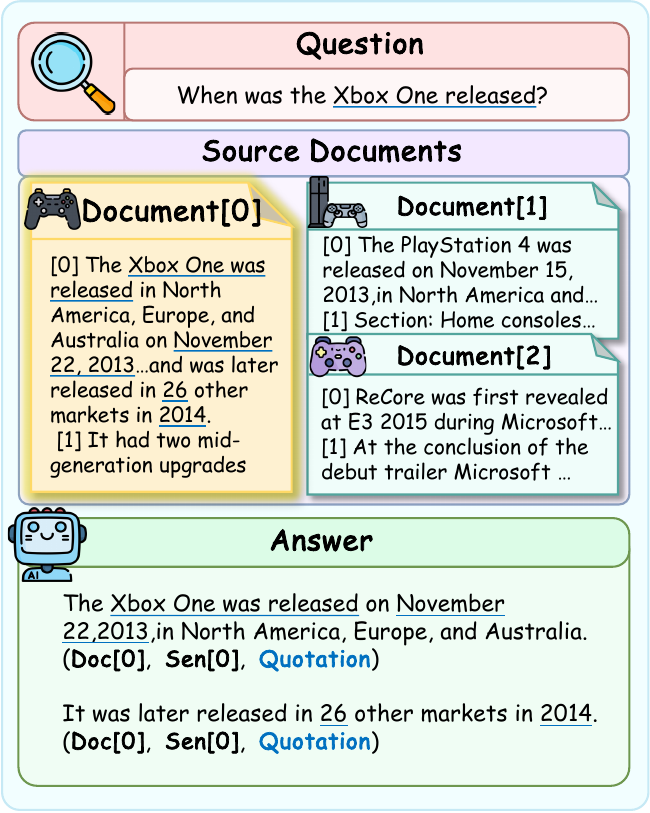}
    \caption{Example ReFInE instance illustrating the \textbf{Quotation} relation type.}
    \label{fig:proveasqa-quotation}
\end{figure}

\begin{figure}[t]
\setlength{\abovecaptionskip}{0.05cm}    
\setlength{\belowcaptionskip}{-0.5cm}
    \centering
    \includegraphics[width=\linewidth]{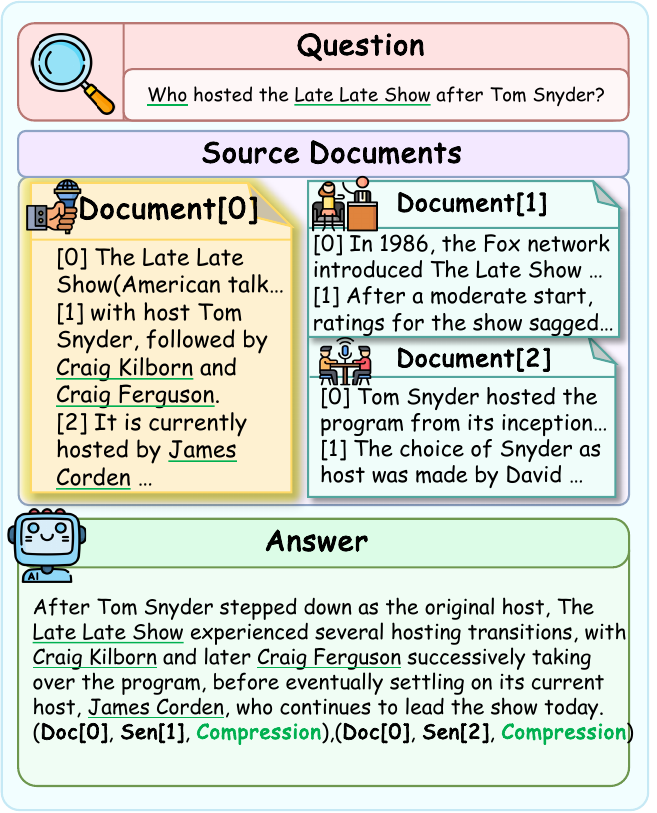}
    \caption{Example ReFInE instance illustrating the \textbf{Compression} relation type.}
    \label{fig:proveasqa-compression}
\end{figure}

\begin{figure}[t]
\setlength{\abovecaptionskip}{0.05cm}    
\setlength{\belowcaptionskip}{-0.5cm}
    \centering
    \includegraphics[width=\linewidth]{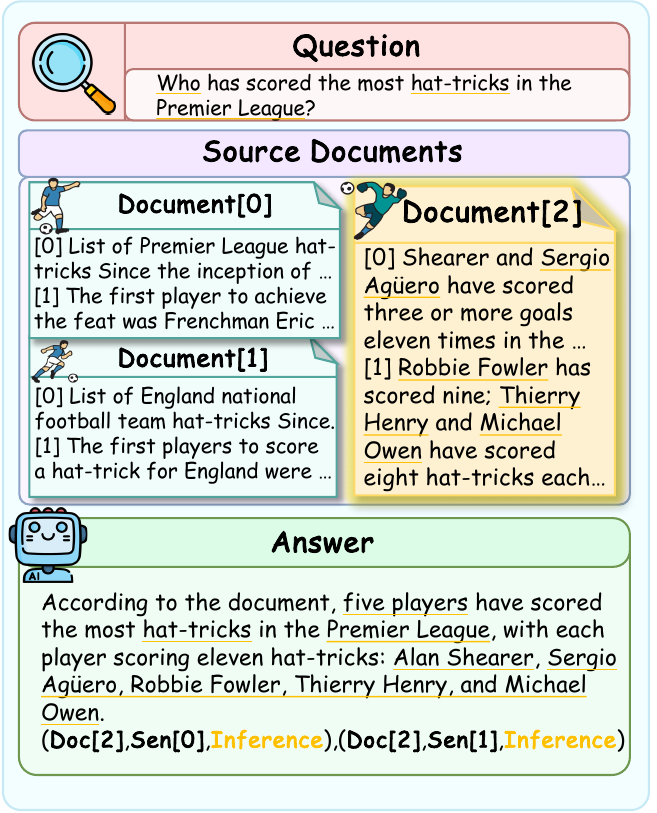}
    \caption{Example ReFInE instance illustrating the \textbf{Inference} relation type.}
    \label{fig:proveasqa-inference}
\end{figure}

To illustrate the final ``message'' format used in ReFInE, we provide examples covering all three provenance relation types. Each instance follows a uniform structure: the \texttt{user} message contains the question together with its associated source documents, and the \texttt{assistant} message provides the answer enriched with sentence-level provenance tags in the \texttt{[PROVE:(doc\_id,\,sent\_id,\,relation)]} format. The examples demonstrate how Quotation, Compression, and Inference relations appear in fully constructed data.

Figure~\ref{fig:proveasqa-quotation} shows a Quotation instance, where each answer sentence closely matches its supporting source sentence. Figure~\ref{fig:proveasqa-compression} illustrates a Compression example in which the answer condenses information across multiple source sentences. Figure~\ref{fig:proveasqa-inference} presents an Inference case where the answer requires multi-sentence or cross-document reasoning grounded in the provided evidence.

\subsection{Additional Dataset Statistics}
\label{app:dataset_stats}

Table~\ref{tab:proveasqa_corpus_stats} reports corpus-level statistics of ReFInE, including provenance density, provenance-triple aggregation, answer length (excluding provenance tags), and source-document granularity.

\begin{table}[t]
  \small
  \centering
  \setlength{\abovecaptionskip}{0.05cm}    
  \setlength{\belowcaptionskip}{-0.5cm}
  \resizebox{\linewidth}{!}{
  \begin{tabular}{l r}
    \toprule
    \textbf{Statistic} & \textbf{Value} \\
    \midrule
    \multicolumn{2}{l}{\textbf{PROVE-tag statistics}} \\
    Total \texttt{[PROVE]} tags & 11{,}467 \\
    \texttt{[PROVE]} tags per answer (avg / median) & 3.96 / 4 \\
    \texttt{[PROVE]} tags per answer (min / max) & 1 / 14 \\
    \midrule
    \multicolumn{2}{l}{\textbf{Provenance-triple statistics}} \\
    Total provenance triples & 22{,}138 \\
    Triples per answer (avg / median) & 7.64 / 7 \\
    Triples per answer (min / max) & 1 / 46 \\
    Triples per [PROVE] tag (avg / median) & 1.98 / 2 \\
    Triples per [PROVE] tag (min / max) & 1 / 18 \\
    \midrule
    \multicolumn{2}{l}{\textbf{Answer length (words, without \texttt{[PROVE]} tags)}} \\
    Answer length (avg / median) & 94.38 / 96 \\
    Answer length (min / max) & 10 / 303 \\
    \midrule
    \multicolumn{2}{l}{\textbf{Source document statistics}} \\
    Sentences per document (avg / median) & 4.52 / 4 \\
    Sentences per document (min / max) & 1 / 19 \\
    Words per document (avg / median) & 88.61 / 97 \\
    Words per document (min / max) & 25 / 103 \\
    \bottomrule
  \end{tabular}
  }
  \caption{Corpus-level statistics of ReFInE, detailing provenance tag density, provenance aggregation, and source document granularity.}
  \label{tab:proveasqa_corpus_stats}
\end{table}

\subsection{Benchmark Comparison Details}
\label{app:benchmark_comparison}

Table~\ref{tab:proveasqa_benchmark_comparison} compares ReFInE with representative benchmarks. We use \ding{51}/\ding{55} to denote Yes/No. ``Avg.\ citations'' denotes the average number of provenance triples in the reference answers. ``Structured tags'' indicates metadata-rich provenance tags (DocID, SentID, Relation) rather than plain indices.

\begin{table*}[t]
  \small
  \centering
  \setlength{\abovecaptionskip}{0.05cm}    
  \setlength{\belowcaptionskip}{-0.1cm}
  \resizebox{\textwidth}{!}{
  \begin{tabular}{l c c c c c c c c}
    \toprule
    \textbf{Benchmark} & \textbf{Year} & \textbf{\#Rel} & \textbf{Avg.\ citations} & \textbf{Sent.-level} & \textbf{Simul.} & \textbf{Generative} & \textbf{Structured tags} & \textbf{Typed relations} \\
    \midrule
    Explicit dataset~\cite{xing-2020-automatic-citation-texts} & 2020 & 1 & 1.00    & \ding{55} & \ding{55} & \ding{51} & \ding{55} & \ding{55} \\
    FEVER~\cite{chen2022gere}                   & 2022 & 3 & 1.86 & \ding{51} & \ding{55} & \ding{55} & \ding{55} & \ding{55} \\
    ASQA~\cite{gao-2023-enabling-llms-citation}                   & 2023 & 1 & 0    & \ding{51} & \ding{51} & \ding{51} & \ding{55} & \ding{55} \\
    WikiRetr~\cite{li-2024-citation-enhanced}                    & 2024 & 2 & 1    & \ding{51} & \ding{55} & \ding{51} & \ding{55} & \ding{55} \\
    SCiFi~\cite{cao2024verifiable}                  & 2024 & 1 & 1.86 & \ding{55} & \ding{51} & \ding{51} & \ding{55} & \ding{55} \\
    ELI5~\cite{cao-2024-calm}                   & 2024 & 1 & 0    & \ding{51} & \ding{55} & \ding{51} & \ding{55} & \ding{55} \\
    FRONT~\cite{huang2024learning}                  & 2024 & 1 & 4.40  & \ding{51} & \ding{51} & \ding{51} & \ding{55} & \ding{55} \\
    MDS~\cite{slobodkin2024attribute}        & 2024 & 1 & 3.00    & \ding{51} & \ding{55} & \ding{51} & \ding{55} & \ding{55} \\
    CG~\cite{li2025scirgc}                 & 2025 & 3 & 1.00    & \ding{51} & \ding{55} & \ding{51} & \ding{55} & \ding{51} \\
    TROVE~\cite{zhu-etal-2025-trove}                  & 2025 & 4 & 1.97 & \ding{51} & \ding{55} & \ding{55} & \ding{51} & \ding{51} \\
    \midrule
    \textbf{ReFInE (Ours)} & -- & 3 & 7.64 & \ding{51} & \ding{51} & \ding{51} & \ding{51} & \ding{51} \\
    \bottomrule
  \end{tabular}
  }
  \caption{Comparison between ReFInE and representative benchmarks. We use \ding{51}/\ding{55} to denote Yes/No. ``Avg.\ citations'' denotes the average number of provenance triples or their equivalents in the reference answers.
 ``Structured tags'' indicates metadata-rich provenance tags (DocID, SentID, Relation) rather than plain indices. ReFInE distinguishes itself by enforcing generation-time, sentence-level provenance with explicit relation typing.}
  \label{tab:proveasqa_benchmark_comparison}
\end{table*}

\section{Additional Experimental Details}
\label{sec:appendix_exp_details}

\subsection{Inference Prompt for Provenance-Aware Generation}
\label{subsec:appendix_inference_prompt}

Figure~\ref{inference_prompt} presents the unified inference prompt used across all models in our experiments. The prompt enforces evidence-based generation conditioned on the provided source documents and requires each factual sentence to be accompanied by a structured provenance tag. It explicitly defines the three provenance relation types—Quotation, Compression, and Inference—and specifies exclusion rules to avoid annotating non-factual content.

\begin{figure*}[t]
\setlength{\abovecaptionskip}{0.05cm}    
\setlength{\belowcaptionskip}{-0.3cm}
\centering
\begin{promptbox}[Inference Prompt]

\scriptsize
You are a rigorous AI assistant specializing in traceable Question Answering. Your task is to generate an accurate, fluent, and factual answer based ONLY on the provided Source Documents.

\textbf{CORE INSTRUCTIONS:}

1. \textbf{Evidence-Based Generation}:  
Every sentence containing factual information must be supported by the Source Documents.

2. \textbf{0-Based Indexing}:  
Always use 0-based indexing for Document IDs (\texttt{Doc[0] → "0"}) and Sentence IDs exactly as they appear in the input.

3. \textbf{Strict Citation Format}:  
Append a citation tag at the end of every factual sentence.  
Format:  
\quad \texttt{[PROVE: ("doc\_id", "sent\_id", "relation")]}  

If multiple sources support the same sentence, merge them inside a \emph{single} PROVE tag:  
Correct:  
\quad \texttt{[PROVE: ("0","1","Quotation"), ("1","3","Inference")]}  
Incorrect:  
\quad \texttt{[PROVE: ("0","1","Quotation")] [PROVE: ("1","3","Inference")]}

---

\textbf{ATTRIBUTION RELATIONS}

Choose exactly one relation for each citation tuple:

- \textbf{Quotation}: The answer sentence partially or fully copies the wording from the source (even with small rewrites).
- \textbf{Compression}: The answer sentence condenses information from one or more source sentences.
- \textbf{Inference}: The answer sentence expresses a fact implied but not explicitly stated by the source.

---

\textbf{EXCLUSION RULES (Do NOT cite):}

• Transition phrases (e.g., “To summarize”, “In conclusion”).  
• Sentences that describe document structure without factual content.

---

\textbf{ONE-SHOT EXAMPLE}

\textbf{User Input:}

\textbf{Question:}  
What was the 2018 All-Ireland Senior Hurling Championship Final?

\textbf{Source Documents:}

\textbf{Doc[0]}:  
[0] "2018 All-Ireland Senior Hurling Championship Final The 2018 All-Ireland Senior Hurling Championship Final, the 131st event of its kind and the culmination of the 2018 All-Ireland Senior Hurling Championship, was played at Croke Park in Dublin on 19 August 2018.[1]  
The final was shown live in Ireland on RTÉ Two ...  
Match commentary was provided by Marty Morrissey ...  
The game was also shown live on Sky Sports ..."

\textbf{Doc[1]}:  
[0] "2018 All-Ireland Minor Hurling Championship ... ended on 19 August 2018 ..."

\textbf{Doc[2]}:  
[0] "2018 All-Ireland Senior Hurling Championship ... 131st staging ..."

---

\textbf{Model Output:}

The 2018 All-Ireland Senior Hurling Championship Final was the 131st event of its kind and the culmination of the 2018 All-Ireland Senior Hurling Championship.  
\texttt{[PROVE: ("0","0","Quotation"), ("2","0","Compression")]}  

It was played at Croke Park in Dublin on August 19, 2018, and was shown live in Ireland on RTÉ Two as part of "The Sunday Game" live program...  
\texttt{[PROVE: ("0","0","Quotation"), ("0","1","Quotation")]}  

Match commentary was provided by Marty Morrissey with analysis by Michael Duignan.  
\texttt{[PROVE: ("0","2","Quotation")]}  

The game was also shown live on Sky Sports, presented by Rachel.  
\texttt{[PROVE: ("0","3","Quotation")]}  

\end{promptbox}
\caption{Prompt used for model inference with structured provenance.}
\label{inference_prompt}
\end{figure*}

\subsection{LLM Judge for Traceable QA}
\label{subsec:appendix_llm_judge}

To achieve a more precise and interpretable evaluation, we decouple LLM-based judging into two independent components: \textbf{text generation quality} and \textbf{traceability quality}.  
Instead of relying on a single judge that conflates linguistic quality with citation correctness, we adopt two specialized LLM judges, each focusing on a distinct evaluation objective.

The first judge evaluates the natural language quality of the generated answer, while the second judge exclusively assesses the correctness and completeness of the provenance annotations.  
This separation enables finer-grained diagnosis of model errors, distinguishing between deficiencies in answer quality and failures in attribution or reasoning.

\noindent\textbf{Text Generation Quality Judge}
Figure~\ref{fig:text_quality_prompt} shows the prompt used to evaluate the textual quality of model responses.  
The judge compares the generated answer against the question, source documents, and ground-truth labels, and assigns a score from 0 to 5 based on accuracy, fluency, and completeness.  
This judge does not consider provenance tags and focuses solely on the quality of the natural language answer.

\begin{figure*}[t]
\centering
\begin{promptbox}[Text Quality Evaluation Prompt]

\scriptsize
You are a content quality evaluation expert. Your task is to evaluate the text quality of a Q\&A model's response.

\textbf{Input Data:}
\begin{itemize}\setlength{\itemsep}{1pt}
  \item \texttt{question}: The user's query.
  \item \texttt{documents}: The source material provided to the model.
  \item \texttt{labels}: The standard reference answer (Ground Truth).
  \item \texttt{response}: The model's generated answer.
\end{itemize}

\textbf{Objective:}  
Evaluate the natural language answer.  
Compare the model's \texttt{response} against the \texttt{labels} and \texttt{documents}.

\textbf{Scoring Criteria (0--5):}  

Focus on \textbf{Accuracy}, \textbf{Fluency}, and \textbf{Completeness}.  

5 (Perfect): Accurate, fluent, complete. No hallucinations.  

4 (Good): Basically accurate. Covers main points.  

3 (Acceptable): Captures core answer, minor slips.  

2 (Poor): Misses key info, hallucinations, or poor grammar.  

1 (Very Poor): Barely relevant or severe errors.  

0 (Useless): Completely wrong or empty.

---

\textbf{Output Format (JSON only):}
\begin{verbatim}
{
  "id": "<id>",
  "question": "<question>",
  "text_quality_score": <integer 0-5>,
  "text_quality_reasoning": "<Concise explanation>"
}
\end{verbatim}

---

\textbf{Reference Example:}

\textbf{Reference input:}
\begin{verbatim}
{
  "id": "e318e8cf-cfa9-4889-8a2e-b37b18b64ac7",
  "question": "What happened to the Milwaukee Brewers in the 2008
  National League Division Series?",
  "documents": { ... },
  "response": "...",
  "labels": "..."
}
\end{verbatim}

\textbf{Reference output:}
\begin{verbatim}
{
  "id": "e318e8cf-cfa9-4889-8a2e-b37b18b64ac7",
  "question": "What happened to the Milwaukee Brewers in the 2008
  National League Division Series?",
  "text_quality_score": 4,
  "text_quality_reasoning":
    "Response accurately and fluently states the Brewers played
     and were eliminated by the Phillies in the 2008 NLDS.
     However, it omits the context of clinching a wild card spot
     with a 90-72 record, making it slightly less complete."
}
\end{verbatim}

\end{promptbox}
\caption{Prompt used for evaluating the text generation quality of model responses.}
\label{fig:text_quality_prompt}
\end{figure*}

\noindent\textbf{Traceability Quality Judge}
Figure~\ref{fig:traceability_prompt} illustrates the prompt used to evaluate provenance correctness. 
This judge focuses exclusively on the \texttt{[PROVE]} tags, verifying both the accuracy and completeness of attribution relationships with respect to the ground-truth labels and source documents.

Crucially, the judge explicitly penalizes missing relationship types required by the labels, enabling principled detection of recall errors in traceability generation.

\begin{figure*}[t]
\centering
\begin{promptbox}[Traceability Evaluation Prompt]

\scriptsize
You are a rigorous citation evaluation expert. Your task is to evaluate the \textbf{Traceability} of a Q\&A model's response.  
You must verify the model's \texttt{[PROVE]} tags against the provided \texttt{documents} and the standard ground-truth \texttt{labels}.

\textbf{Input Data:}
\begin{itemize}\setlength{\itemsep}{1pt}
  \item \texttt{id}: Data item ID.
  \item \texttt{question}: The user's query.
  \item \texttt{documents}: The source material.
  \item \texttt{labels}: The standard reference answer (Ground Truth).
  \item \texttt{response}: The model's generated answer (containing \texttt{[PROVE]} tags).
\end{itemize}

The definitions of attribution relationships in the \texttt{[PROVE]} tags are as follows:  

Quotation: The answer sentence partially or fully copies sentences from the source document.  

Compression: The answer sentence condenses information from one or more sentences.  

Inference: The answer sentence is based on information implied by the source document.

Evaluate the correctness and completeness of the \texttt{[PROVE]} tags by comparing the \texttt{response} against the \texttt{labels}.

\textbf{CRITICAL SCORING LOGIC:}  

1. Check for Missing Types (Recall): If a relationship type exists in \texttt{labels} but is \textbf{NOT} in \texttt{response}, the score for that type \textbf{MUST} be \textbf{0}.  

2. Check for Unused Types: If a relationship type is \textbf{NOT} in \texttt{labels} AND \textbf{NOT} in \texttt{response}, the score \textbf{MUST} be \texttt{null}.  

3. Check for Accuracy (Precision): If the type exists in \texttt{response}, score it based on correctness (0--5) relative to the documents:  

5: Most tags for this type are correct and match the standard labels logic.  

4: More than half for this type are correct; minor issues with boundaries.  

3: Mixed accuracy; only about one-third for this type are correct.  

2: Only about one-eighth of them are correct.  

1: Few tags are correct; vast majority are hallucinations.  

0: All tags are hallucinations, irrelevant, OR the type is required by labels but missing in response.

\textbf{2. Overall Citation Score (0--5)}  

Provide a holistic score for the model's citation performance.  

5: Perfect. Captures all relationships required by \texttt{labels} with accurate citations.  

4: Good. Captured all required relationships but with minor inaccuracies in index or boundaries.  

3: Acceptable but flawed. Missed one relationship type or has several inaccurate citations.  

2: Poor. Missed multiple required relationships or most citations are wrong.  

1: Very Poor. Citations are mostly hallucinated or irrelevant.  

0: No valid citations or complete failure to follow instructions.

\textbf{Special Note on Indexing:}  

Source \texttt{documents} are 0-indexed. If the answer text refers to ``Document 1'' but the \texttt{[PROVE]} tag uses index \texttt{"0"}, this is considered correct and should not be penalized.

\textbf{Output Format:}  

Strictly output in the following JSON format. Do not add any extra explanations.

\begin{verbatim}
{
  "id": "<id of the current data item>",
  "question": "<question of the current data item>",
  "relationship_scores": {
    "Quotation": <integer from 0 to 5 OR null>,
    "Compression": <integer from 0 to 5 OR null>,
    "Inference": <integer from 0 to 5 OR null>
  },
  "overall_citation_score": <integer from 0 to 5>,
  "citation_reasoning":
    "<Explain the scores. Explicitly mention if a type required
      by 'labels' was missed (Recall error) or if the generated
      tags were inaccurate (Precision error).>"
}
\end{verbatim}

\textbf{Reference Example (abridged):}

\begin{verbatim}
Reference input: {...}

Reference output:
{
  "relationship_scores": {
    "Quotation": 3,
    "Compression": null,
    "Inference": 0
  },
  "overall_citation_score": 3,
  "citation_reasoning":
    "Recall error: the required 'Inference' relationship is missing.
     Precision error: one 'Quotation' tag is misaligned with the
     referenced document sentence."
}
\end{verbatim}

\end{promptbox}
\caption{Prompt used for evaluating the correctness and completeness of provenance annotations (\texttt{[PROVE]} tags), with explicit modeling of recall and precision errors.}
\label{fig:traceability_prompt}
\end{figure*}

\subsection{Human Evaluation Guidelines}
\label{subsec:appendix_human_eval}

We conduct human evaluation to complement automatic and LLM-based metrics. Three expert annotators with prior experience in question answering and evidence annotation participate in the study. We randomly sample 200 evaluation instances from the test set and collect ratings for each model output. Each instance is independently evaluated by all annotators, and final scores are obtained by averaging across raters.

\paragraph{Answer quality (0--5).}
Raters score the natural-language answer while ignoring all provenance tags. Scores reflect accuracy with respect to the reference answer and the source documents, as well as fluency and completeness. A score of 5 indicates a fully correct and fluent answer with no substantive omissions or errors, while 0 indicates an unusable response (e.g., empty, refusal, or entirely incorrect).

\paragraph{Provenance quality (0--5).}
Raters assess whether provenance tags correctly and sufficiently support the answer. Relation types follow the same definitions as in our task (Quotation, Compression, Inference). We apply two key rules for relation-type scoring: (i) if a relation type appears in the reference but is missing from the model output, the score for that type is 0; (ii) if a relation type appears in neither the reference nor the model output, the score for that type is null. When a relation type is used by the model, raters assign a 0--5 score based on correctness of the cited evidence and relation typing. Raters also provide an overall provenance score (0--5) that summarizes citation correctness and coverage. We do not penalize purely textual mentions of 1-based document numbering if the provenance tags correctly map to the underlying 0-based document identifiers.

\subsection{LLM-as-a-Judge Breakdown}
\label{subsec:appendix_llm_judge_table}

\begin{table*}[t]
  \centering
  \setlength{\abovecaptionskip}{0.05cm}    
  \setlength{\belowcaptionskip}{-0.5cm}
    \resizebox{\textwidth}{!}{
    \begin{tabular}{lcccccc}
    \toprule
    \textbf{Model} &
    \textbf{Text Quality} &
    \textbf{Prov.} &
    \textbf{Quotation} &
    \textbf{Compression} &
    \textbf{Inference} &
    \textbf{Avg} \\
    \midrule
    \multicolumn{7}{c}{Open-Source} \\
    \midrule
    \multicolumn{1}{l|}{Baichuan2-7B~\cite{yang2023baichuan}} & 3.04  & 0.37  & 0.39  & 0.05  & 0.05  & 0.78  \\
    \multicolumn{1}{l|}{Vicuna-7b-v1.5~\cite{zheng2023judging}} & 3.24  & 0.84  & 0.86  & 0.35  & 0.19  & 1.10  \\
    \multicolumn{1}{l|}{InternLM2.5-7B~\cite{cai2024internlm2}} & 3.83  & 1.69  & 1.71  & 1.10  & 0.03  & 1.67  \\
    \multicolumn{1}{l|}{Hunyuan-7B~\cite{zheng2025hunyuan}} & 3.05  & 2.11  & 2.64  & 0.09  & 0.69  & 1.72  \\
    \multicolumn{1}{l|}{Yi-1.5-9B~\cite{liu2025open}} & 3.84  & 2.12  & 2.62  & 0.23  & 0.17  & 1.80  \\
    \multicolumn{1}{l|}{Llama-3.1-8B~\cite{dubey2024llama}} & 3.78  & 2.20  & 2.56  & 1.05  & 0.42  & 2.00  \\
    \multicolumn{1}{l|}{GLM-4-9B~\cite{glm2024chatglm}} & 4.00  & 2.67  & 3.51  & 0.14  & 0.48  & 2.16  \\
    \multicolumn{1}{l|}{Qwen3-8B~\cite{yang2025qwen3}} & 4.02  & 2.67  & 3.42  & 0.66  & 0.47  & 2.25  \\
    \multicolumn{1}{l|}{Gemma-3-12B~\cite{team2025gemma}} & 3.97  & 2.89  & 4.03  & 1.06  & 0.42  & 2.47  \\
    \multicolumn{1}{l|}{Qwen3-14B~\cite{yang2025qwen3}} & 4.10  & \cellcolor[rgb]{ .933,  .973,  .91}3.03  & 4.06  & \cellcolor[rgb]{ .933,  .973,  .91}1.56  & 0.21  & 2.59  \\
    \multicolumn{1}{l|}{GLM-4.5-355B~\cite{glm2024chatglm}} & \cellcolor[rgb]{ .992,  .941,  .902}\textbf{4.29 } & 3.02  & 4.13  & 0.91  & \cellcolor[rgb]{ .933,  .973,  .91}0.79  & \cellcolor[rgb]{ .933,  .973,  .91}2.63  \\
    \midrule
    \multicolumn{7}{c}{Closed-Source} \\
    \midrule
    \multicolumn{1}{l|}{Kimi~\cite{team2025kimi}} & 4.13  & 2.70  & 3.60  & 0.44  & 0.13  & 2.20  \\
    \multicolumn{1}{l|}{GPT-5~\cite{achiam2023gpt}} & 4.14  & 2.34  & 2.39  & 1.97  & 0.29  & 2.23  \\
    \multicolumn{1}{l|}{Gemini 2.5 Pro~\cite{comanici2025gemini}} & \cellcolor[rgb]{ .933,  .973,  .91}4.24  & 3.02  & \cellcolor[rgb]{ .933,  .973,  .91}4.33  & 1.09  & 0.19  & 2.57  \\
    \midrule
    \multicolumn{7}{c}{Ours} \\
    \midrule\multicolumn{1}{l}{GenProve} & 3.98  & \cellcolor[rgb]{ .992,  .941,  .902}\textbf{3.42 } & \cellcolor[rgb]{ .992,  .941,  .902}\textbf{4.43 } & \cellcolor[rgb]{ .992,  .941,  .902}\textbf{2.45 } & \cellcolor[rgb]{ .992,  .941,  .902}\textbf{1.40 } & \cellcolor[rgb]{ .933,  .973,  .91}\textbf{3.14 } \\
    \bottomrule
    \end{tabular}%
    }
  \caption{Breakdown of LLM-as-a-Judge evaluation. GenProve achieves superior overall ratings primarily through significant gains in complex \textit{Compression} and \textit{Inference} relations.}
  \label{tab:llm_judge_breakdown}%
\end{table*}%

Table~\ref{tab:llm_judge_breakdown} reports the LLM-as-a-judge breakdown on ReFInE. 
The \textbf{Avg} column matches the LLM-as-judge score reported in the main results, while \textbf{Text Quality} and the provenance columns explain where that end-to-end score comes from. 
This decomposition directly aligns with our motivation: for trustworthy generation, high-level answer fluency is insufficient unless each sentence is supported by correctly localized evidence and an appropriate relation type.

\textbf{Overall}, the table shows that end-to-end differences are driven mainly by provenance rather than surface answer quality. 
For competitive systems, Text Quality scores concentrate in a relatively narrow range, whereas Overall Prov.\ and relation-specific provenance scores vary substantially. 
As a result, models with comparable Text Quality can still diverge sharply in Avg, which indicates that traceability correctness is the primary bottleneck captured by this benchmark.

\textbf{Across model groups}, open-source systems exhibit the largest dispersion. 
Earlier baselines such as Baichuan2-7B~\cite{yang2023baichuan} and Vicuna-7B-v1.5~\cite{zheng2023judging} show low provenance scores, consistent with the weaker schema-following and attribution behavior observed in the main results. 
Stronger open-source instruction-tuned models, including Qwen3~\cite{yang2025qwen3} and GLM-4~\cite{glm2024chatglm}, achieve much higher provenance scores and therefore higher Avg. 
Among non-GenProve systems, GLM-4.5~\cite{glm2024chatglm} provides the strongest overall baseline by Avg and overall provenance. 
Closed-source models are competitive: Gemini~2.5~Pro~\cite{comanici2025gemini} yields the strongest closed-source Avg and provenance, while GPT-5~\cite{achiam2023gpt} and Kimi~\cite{team2025kimi} show weaker provenance breakdowns despite strong text scores.

\textbf{From the metric perspective}, the relation-specific columns reveal a stable difficulty pattern. 
Quotation receives the highest scores for most models, which indicates that direct reuse attribution is relatively easy to judge and to satisfy. 
In contrast, Compression and Inference scores remain low for many systems, even when their Quotation scores are strong, and these two relations largely determine differences in Overall Prov.\ and thus Avg. 
This breakdown clarifies that improvements in Avg mainly coincide with better handling of abstraction and reasoning-based provenance, rather than with changes in Text Quality alone.

\subsection{Full Ablation Results}
\label{subsec:appendix_ablation_full}

\begin{table*}[t]
  \small
  \setlength{\abovecaptionskip}{0.05cm}    
  \setlength{\belowcaptionskip}{-0.5cm}
  \centering
  \resizebox{\linewidth}{!}{
  \begin{tabular}{lcccccccccc}
    \toprule
    \textbf{Model} &
    \textbf{ROUGE-L} & \textbf{BLEU} & \textbf{METEOR} & \textbf{BERTScore} & \textbf{MoverScore} &
    \textbf{Prec.} & \textbf{Rec.} & \textbf{F1} &
    \textbf{Format} & \textbf{Judge} \\
    \midrule
    GenProve & 57.25 & 42.22 & 59.39 & 61.98 & 51.04 & 54.96 & 51.26 & 51.21 & 99.85 & 3.14 \\
    w/o Provenance Reward & 36.85 & 11.94 & 25.06 & 46.96 & 30.04 & 27.10 & 24.08 & 24.67 & 96.66 & 2.20 \\
    w/o Similarity Reward & 38.72 & 24.60 & 54.97 & 44.50 & 35.49 & 67.27 & 58.60 & 60.32 & 95.74 & 2.71 \\
    w/o GRPO & 56.08 & 41.82 & 59.96 & 60.70 & 49.79 & 53.22 & 51.65 & 50.48 & 99.70 & 2.62 \\
    \bottomrule
  \end{tabular}
  }
  \caption{Full ablation results on ReFInE (EVAL).}
  \label{tab:ablation_full}
\end{table*}

Table~\ref{tab:ablation_full} clarifies how each training component affects different dimensions of performance. GRPO primarily improves end-to-end utility under joint requirements, as reflected by a higher judge score compared with the SFT-only model. The provenance reward directly strengthens sentence-level attribution, and its removal leads to a broad collapse in provenance precision, recall, and F1. In contrast, the similarity reward provides an explicit pressure toward content faithfulness and wording alignment; removing it yields high provenance F1 but reduces answer-quality metrics and lowers the judge score, which suggests that provenance matching alone can be satisfied by outputs that are less faithful to the reference answer. The full model balances these pressures and achieves the best overall trade-off, which matches the intended objective of generation-time fine-grained provenance: accurate answers with localized and correctly typed evidence.

\subsection{Human Evaluation Results}
\label{subsec:appendix_human_results}

\begin{table*}[t]
  \centering
  \setlength{\abovecaptionskip}{0.05cm}    
  \setlength{\belowcaptionskip}{-0.5cm}
  \resizebox{\textwidth}{!}{
    \begin{tabular}{ccccccc}
    \toprule
    \textbf{Model} &
    \textbf{Text Quality} &
    \textbf{Prov.} &
    \textbf{Quotation} &
    \textbf{Compression} &
    \textbf{Inference} &
    \textbf{Avg} \\
    \midrule
    \multicolumn{7}{c}{Open-Source} \\
    \midrule
    \multicolumn{1}{l|}{Baichuan2-7B~\cite{yang2023baichuan}} & 3.65  & 0.41  & 0.51  & 0.20  & 0.10  & 0.98  \\
    \multicolumn{1}{l|}{Vicuna-7b-v1.5~\cite{zheng2023judging}} & 2.82  & 0.62  & 0.75  & 0.16  & 0.41  & 0.95  \\
    \multicolumn{1}{l|}{InternLM2.5-7B~\cite{cai2024internlm2}} & 3.91  & 1.49  & 1.51  & 1.16  & 0.10  & 1.64  \\
    \multicolumn{1}{l|}{Hunyuan-7B~\cite{zheng2025hunyuan}} & 3.45  & 2.38  & 3.12  & 0.61  & 1.08  & 2.13  \\
    \multicolumn{1}{l|}{Yi-1.5-9B~\cite{liu2025open}} & 3.98  & 2.05  & 2.44  & 0.10  & 0.20  & 1.76  \\
    \multicolumn{1}{l|}{Llama-3.1-8B~\cite{dubey2024llama}} & 3.82  & 2.40  & 2.74  & 1.16  & 1.20  & 2.27  \\
    \multicolumn{1}{l|}{GLM-4-9B~\cite{glm2024chatglm}} & 4.02  & 2.45  & 2.85  & 0.33  & 0.92  & 2.11  \\
    \multicolumn{1}{l|}{Qwen3-8B~\cite{yang2025qwen3}} & 4.03  & 2.63  & 3.36  & 0.82  & \cellcolor[rgb]{ .918,  .98,  .945}1.26  & 2.42  \\
    \multicolumn{1}{l|}{Gemma-3-12B~\cite{team2025gemma}} & 3.90  & 3.03  & 3.69  & 1.14  & 0.96  & 2.54  \\
    \multicolumn{1}{l|}{Qwen3-14B~\cite{yang2025qwen3}} & 3.99  & 2.91  & 3.25  & 1.20  & 0.47  & 2.36  \\
    \multicolumn{1}{l|}{GLM-4.5-355B~\cite{glm2024chatglm}} & 4.15  & 2.90  & 3.59  & 0.99  & 0.30  & 2.39  \\
    \midrule
    \multicolumn{7}{l}{Closed-Source} \\
    \midrule
    \multicolumn{1}{l|}{Kimi~\cite{team2025kimi}} & 4.21  & 2.44  & 3.85  & 0.38  & 0.21  & 2.22  \\
    \multicolumn{1}{l|}{GPT-5~\cite{achiam2023gpt}} & 4.24  & 3.17  & 3.22  & \cellcolor[rgb]{ .918,  .98,  .945}2.33  & 0.63  & \cellcolor[rgb]{ .918,  .98,  .945}2.72  \\
    \multicolumn{1}{l|}{Gemini 2.5 Pro~\cite{comanici2025gemini}} & \cellcolor[rgb]{ 1,  .914,  .91}\textbf{4.36 } & \cellcolor[rgb]{ .918,  .98,  .945}3.23  & \cellcolor[rgb]{ .918,  .98,  .945}3.95  & 1.28  & 0.16  & 2.60  \\
    \midrule
    \multicolumn{7}{c}{Ours} \\
    \midrule\multicolumn{1}{l}{GenProve}
     & \cellcolor[rgb]{ .918,  .98,  .945}4.29  & \cellcolor[rgb]{ 1,  .914,  .91}\textbf{3.83 } & \cellcolor[rgb]{ 1,  .914,  .91}\textbf{4.16 } & \cellcolor[rgb]{ 1,  .914,  .91}\textbf{2.57 } & \cellcolor[rgb]{ 1,  .914,  .91}\textbf{1.94 } & \cellcolor[rgb]{ 1,  .914,  .91}\textbf{3.36 } \\
    \bottomrule
    \end{tabular}%
    }
  \caption{Human evaluation results. The manual ratings corroborate automatic metrics, confirming GenProve's superiority in generating verifiable answers with accurate fine-grained provenance.}
  \label{tab:human_scores}
\end{table*}%
We conduct human evaluation to validate the judge-based results and to provide a fine-grained view of answer quality and provenance quality. For each model, raters score (i) answer quality while ignoring provenance tags and (ii) provenance quality, including an overall provenance score and relation-specific provenance scores for Quotation, Compression, and Inference. Table~\ref{tab:human_scores} reports the averaged scores across models.

\noindent\textbf{Overall.}
Human scores broadly track the main results: models that achieve higher judge scores also receive higher overall human scores (Avg), which supports the use of judge-based evaluation for this task.

\noindent\textbf{Model groups.}
Open-source models show a wide spread in provenance-related scores, ranging from near-failing provenance (e.g., low Prov. and relation scores) to much stronger traceability among recent instruction-tuned systems. Closed-source models are generally strong on answer quality, while provenance remains uneven across relation types. GenProve achieves the highest Avg and the strongest overall provenance score, indicating that improvements are not limited to fluency but extend to evidence attribution.

\noindent\textbf{Metric dimensions.}
Answer scores are relatively high for many models, while provenance scores are substantially lower and vary more by relation type. In particular, Quotation tends to score higher than Compression and Inference, consistent with the increasing difficulty of abstraction and reasoning under sentence-level evidence constraints. GenProve improves all three relation types and shows especially large gains on Compression and Inference, which aligns with the goal of generation-time fine-grained provenance.

\section{Error Analysis}
\label{app:error_analysis}

We provide additional qualitative error analysis to illustrate representative failure modes in sentence-level provenance generation.
Although GenProve substantially improves attribution accuracy, generation-time provenance supervision remains challenging in several other respects.
The following examples highlight three common error patterns observed across models: unsynchronized provenance generation, incomplete provenance coverage, and incorrect provenance localization.

\begin{figure}[t]
  \centering
  \includegraphics[width=\linewidth]{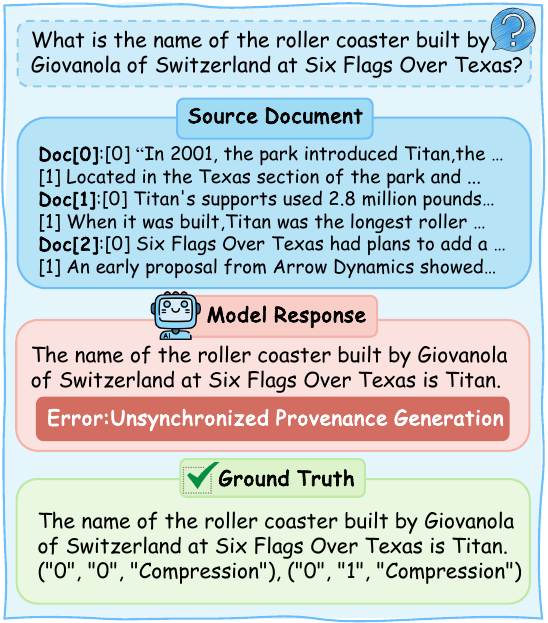}
  \caption{Unsynchronized provenance generation, where answer content and provenance tags become misaligned during decoding.}
  \label{fig:error_unsync}
  \vspace{-2mm}
\end{figure}

\paragraph{Unsynchronized Provenance Generation.}
As shown in Figure~\ref{fig:error_unsync}, the model may produce answer content and provenance tags in an unsynchronized manner.
Provenance annotations are delayed or structurally detached from the sentences they are intended to support, resulting in outputs that are only partially traceable.
This pattern reflects the difficulty of tightly coupling natural language generation with sentence-level attribution decisions during decoding.

\begin{figure}[t]
  \centering
  \includegraphics[width=\linewidth]{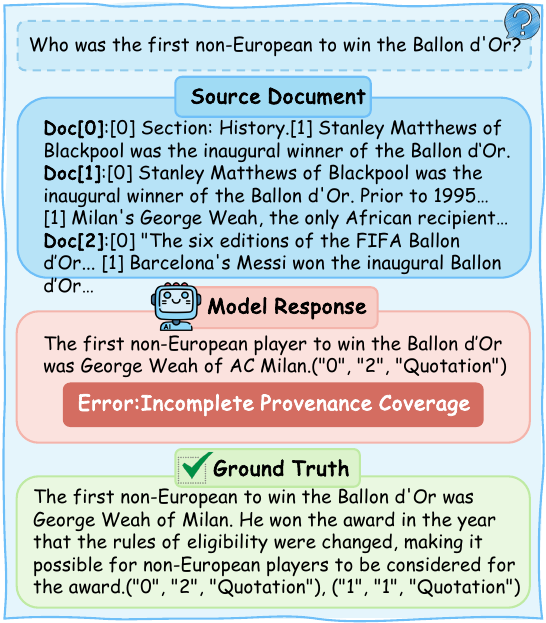}
  \caption{Incomplete provenance coverage, where some factual sentences are not accompanied by the required provenance tags.}
  \label{fig:error_incomplete}
  \vspace{-2mm}
\end{figure}

\paragraph{Incomplete Provenance Coverage.}
Figure~\ref{fig:error_incomplete} illustrates a recall failure in provenance generation: the model produces factually plausible answer sentences but omits provenance tags for some of them.
Such errors break sentence-level verifiability even when the content itself is supported by the source documents.
This pattern suggests that the model underestimates the need for explicit attribution under strict coverage requirements.

\begin{figure}[t]
  \centering
  \includegraphics[width=\linewidth]{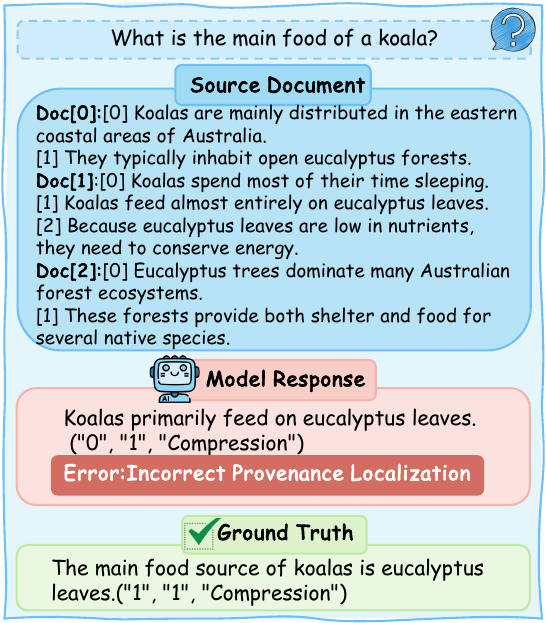}
  \caption{Incorrect provenance localization, where the cited document or sentence indices do not actually support the generated claim.}
  \label{fig:error_localization}
  \vspace{-2mm}
\end{figure}

\paragraph{Incorrect Provenance Localization.}
As shown in Figure~\ref{fig:error_localization}, the model may emit well-formed provenance tags, but the referenced document or sentence indices do not actually contain the supporting evidence.
Although the cited source is often topically related, the precise sentence-level grounding is incorrect.
This error highlights the challenge of fine-grained evidence localization in multi-document settings, where topical relevance alone is insufficient for exact provenance tracing.

\section{Potential Risks}

While generation-time provenance improves transparency and auditability, it also introduces several potential risks that merit careful consideration.

\paragraph{Over-reliance on provenance signals.}
Typed, sentence-level provenance may give users a strong sense of trust in generated answers. However, correct provenance does not guarantee that a statement is fully accurate or appropriate for a given context. Provenance should therefore be interpreted as an aid for inspection rather than a definitive validation of correctness, especially in high-stakes domains.

\paragraph{False sense of completeness.}
Our framework focuses on identifying supporting evidence for generated sentences, but it does not ensure that all relevant evidence has been considered. A model may provide plausible and correctly typed provenance while still omitting counter-evidence or alternative interpretations present in the source documents.

\paragraph{Annotation and evaluation bias.}
PROVE-ASQA relies on LLM-assisted annotation followed by expert validation. Although we apply multi-stage screening and human verification, residual biases from annotator judgment or model priors may affect relation labeling, particularly for subjective cases such as \emph{Inference}. These biases could influence both training and evaluation outcomes.

\paragraph{Computational and deployment considerations.}
Generating sentence-level provenance alongside answers increases output length and computational cost, which may limit applicability in latency-sensitive or resource-constrained settings. Careful system design is required to balance transparency with efficiency.

We emphasize that GenProve is intended as a research step toward more accountable text generation. It should be deployed as part of broader human-in-the-loop or verification workflows rather than as a standalone authority on factual correctness.

\end{document}